%% file: 0_main.tex
\documentclass[manuscript,screen]{acmart}
\usepackage{graphicx} 
\usepackage{threeparttable}
\usepackage{tabularx,booktabs}
\usepackage{makecell}
\usepackage{multirow}
\usepackage{amsmath,amsfonts}
\usepackage{algorithm}  
\usepackage{algorithmicx}  
\usepackage{algpseudocode}
\usepackage{subcaption}
\usepackage{bbding}
\renewcommand\footnotetextcopyrightpermission[1]{}
\AtBeginDocument{%
  }

\begin{document}

\title{TCL: Enabling Fast and Efficient Cross-Hardware Tensor Program Optimization via Continual Learning}

\author{Chaoyao Shen}
\email{shen_chaoyao@seu.edu.cn}
\orcid{0009-0008-3779-5659}
\author{Linfeng Jiang}
\email{linfengjiang@seu.edu.cn}
\orcid{0009-0005-2807-4719}
\affiliation{%
  \institution{Southeast University}
  \city{Nanjing}
  \state{Jiangsu}
  \country{China}
}

\author{Yixian Shen}
\orcid{0000-0001-8447-872X}
\affiliation{%
  \institution{University of Amsterdam}
  \city{Amsterdam}
  \country{Netherlands}}
   \email{y.shen@uva.nl}

\author{Tao Xu}
\orcid{0009-0008-4528-6117}
\affiliation{%
  \institution{Southeast University}
  \city{Nanjing}
  \state{Jiangsu}
  \country{China}}
  \email{xu_tao@seu.edu.cn}

\author{Guoqing Li}
\orcid{0000-0003-2075-8583}
\authornote{Corresponding author.}
\affiliation{%
  \institution{Southeast University \& Shandong Yunhai Guochuang Innovative Technology Co., Ltd.}
  \city{Nanjing, Jinan}
  \state{Jiangsu, Shandong}
  \country{China}}
  \email{li_guoqing@aa.seu.edu.cn}

\author{Anuj Pathania}
\orcid{0000-0002-5813-7021}
\affiliation{%
  \institution{University of Amsterdam}
  \city{Amsterdam}
  \country{Netherlands}}
  \email{a.pathania@uva.nl}

\author{Andy D. Pimentel}
\orcid{0000-0002-2043-4469}
\affiliation{%
  \institution{University of Amsterdam}
  \city{Amsterdam}
  \country{Netherlands}}
\email{a.d.pimentel@uva.nl}

\author{Meng Zhang}
\orcid{0000-0003-2188-8195}
\authornotemark[1]
\affiliation{%
  \institution{Southeast University}
  \city{Nanjing}
  \state{Jiangsu}
  \country{China}}
\email{zmeng@seu.edu.cn}
\renewcommand{\shortauthors}{C. Shen et al.}

\thanks{New Paper, Not an extension of a Conference Paper}

\begin{abstract}
Deep learning (DL) compilers rely on cost models and auto-tuning to optimize tensor programs for target hardware. 
However, existing approaches depend on large offline datasets, incurring high collection costs and offering suboptimal transferability across platforms.
In this paper, we introduce \textbf{TCL}, a novel efficient and transferable compiler framework for fast tensor program optimization across diverse hardware platforms to address these challenges. 
Specifically, \textbf{TCL} is built on three core enablers:
(1) the \textbf{RDU Sampler}, a data-efficient active learning strategy that selects only 10\% of tensor programs by jointly optimizing \underline{R}epresentativeness, \underline{D}iversity, and \underline{U}ncertainty, substantially reducing data collection costs while maintaining near-original model accuracy
(2) a \textbf{new Mamba-based cost model} that efficiently captures long-range schedule dependencies while achieving a favorable trade-off between prediction accuracy and computational cost through reduced parameterization and lightweight sequence modeling;
and (3)  a \textbf{continuous knowledge distillation framework} that effectively and progressively transfers knowledge across multiple hardware platforms while avoiding the parameter explosion and data dependency issues typically caused by traditional multi-task learning.
Extensive experiments validate the effectiveness of each individual enabler and the holistic TCL framework. 
When optimizing a range of mainstream DL models on both CPU and GPU platforms, TCL achieves, on average, $16.8\times$ and $12.48\times$ faster tuning time, and $1.20\times$ and $1.13\times$ lower inference latency, respectively, compared to Tenset-MLP.
\end{abstract}




\maketitle

\input{1_introduction}
\input{2_Related_works}

\input{3_Overview}

\input{4_Sampling}
\input{5_MambaCost}

\input{6_Knowledge}
\input{7_Experimental}

\input{8_Conclusion}

\bibliographystyle{acm}
\bibliography{my}

\end{document}

%% file: 1_introduction.tex
\section{Introduction}
Deep neural networks (DNNs) have achieved remarkable success in domains such as computer vision~\cite{shenefficient,zhu2025interactive,gourdoumanis2026multi,shen2025ssh,wang2025reasoning,huang2024novel}, natural language processing~\cite{zhang2025neuroada,shen2025macp,shen2025altgen,shen2025ssh}, and autonomous driving~\cite{tampuu2020survey,bi2025adadcp}, driving rapid innovation in AI hardware accelerators~\cite{aghapour2024piqi}. 
Specialized platforms such as NVIDIA GPUs~\cite{foley2017ultra}, Google TPUs~\cite{jouppi2017datacenter}, and GraphCore IPUs~\cite{jia2019dissecting}  meet the growing computational demands of these models.
Deep learning (DL) compilers~\cite{guo2024easter} have emerged as a critical bridge between high-level DNN models and low-level hardware execution.
Traditionally, tensor operators are implemented using manually optimized kernel libraries, such as NVIDIA cuDNN~\cite{chetlur2014cudnn} and Intel OneDNN~\cite{onednn}, which are finely tuned for specific hardware.
While these libraries offer excellent performance, they are often tightly coupled to particular architectures, making them difficult to adapt to new or diverse platforms.
Moreover, they struggle to support emerging or non-standard operators, limiting their applicability in rapidly evolving model architectures.
To address the limitations, search-based DL compilers~\cite{chen2018tvm, baghdadi2019tiramisu} have been developed to automatically generate and optimize tensor programs across a variety of hardware backends.
Unlike static kernel libraries, these compilers explore the space of schedule transformations and select implementations tailored to the target device.
A key component of this process is the \textit{cost model}, which predicts the performance (e.g., latency or throughput) of candidate tensor programs and guides the search process effectively.

Current search-based DL compilers often train cost models online~\cite{ahn2020chameleon, zheng2020flextensor}, by iteratively sampling candidate tensor programs, evaluating their hardware performance, and updating the model during auto-tuning. 
While this dynamic feedback loop enables hardware-specific adaptation, 
it comes at a high cost: each program must be compiled and executed during search, which significantly increases tuning time and demands extensive computing resources.
The release of large-scale tensor program datasets, such as Tenset~\cite{zheng2021tenset}, has made it feasible to shift cost model training offline. 
These datasets provide millions of tensor programs with performance measurements across hardware backends, enabling the development of predictive models without requiring real-time hardware profiling.
Motivated by this, recent work has increasingly focused on offline training of cost models~\cite{steiner2021value}. 
In this paradigm, the cost model is trained once on pre-collected data from a specific hardware platform and can then be used to predict the performance of unseen tensor programs during auto-tuning.
To further reduce hardware-specific data collection, researchers have explored transfer learning strategies~\cite{gibson2022transfer, ryu2021metatune}, which adapt cost models pre-trained on one or more source platforms to new target devices by leveraging shared compiler optimization patterns and applying platform-specific fine-tuning.
This combination of offline training and transfer learning reduces tuning time and resource usage while improving compiler adaptability across diverse hardware environments.

Although DL compilers equipped with offline-trained cost models demonstrate strong performance in terms of tuning quality and efficiency, several practical and architectural challenges remain.
First, collecting a sufficiently large set of tensor programs for each hardware platform is highly time-consuming. In our measurements, data collection on an Intel i7-12700F CPU requires approximately 40 days, while the same process on an NVIDIA GeForce RTX 3080Ti GPU takes over 60 days, posing a major bottleneck for scalability.
Second, the architecture of the cost model fundamentally limits the accuracy and efficiency of performance prediction. Early approaches relied on XGBoost~\cite{chen2018tvm} and MLPs~\cite{zheng2021tenset}, which have since evolved to more expressive models such as LSTMs~\cite{baghdadi2021deep} and Transformers~\cite{ryu2022one}. However, LSTMs suffer from limited parallelism and slow convergence, while Transformers, despite their ability to capture long-range dependencies, impose high computational and memory overhead when applied to long sequences of schedule tokens. These architectural limitations prompted us to design a cost model that can efficiently capture sequence dependencies while maintaining a lightweight approach.
Third, while transfer learning has emerged as a popular technique for enabling cross-platform generalization~\cite{zheng2021tenset, zhao2023moses}, most existing methods are restricted to one-to-one adaptation: knowledge is transferred from a single source hardware to a single target hardware. This restricts their ability to accumulate generalized scheduling knowledge across diverse platforms.
Recent work such as MTL-TLP~\cite{zhai2023tlp} attempts to overcome this limitation using multi-task learning to jointly learn from multiple source platforms. 
While promising, this approach introduces significant scalability issues: the number of trainable parameters and associated computational cost grow proportionally with the number of hardware targets, and the framework requires simultaneous access to data from all source platforms during training, limiting practicality. These limitations have prompted us to develop more flexible multi-platform learning frameworks to efficiently utilize knowledge from multiple source platforms without introducing excessive computational costs.

To address the challenges of existing DL compilers across hardware platforms, namely the high cost of data collection, inefficiencies in cost model design, and suboptimal generalization across hardware,
we propose \textbf{TCL}, a fast, data-efficient, and transferable tensor compiler framework.
At the data level, it incorporates a principled active learning strategy, the \textbf{RDU Sampler}, which selects the most informative tensor programs by jointly considering representativeness, diversity, and uncertainty. This dramatically reduces the number of performance measurements required, achieving competitive or even superior tuning quality with as little as 10\% of the data.
To enhance prediction efficiency, TCL employs a new cost model architecture that is both lightweight and expressive, built upon the \textbf{Mamba Block}~\cite{gu2023mamba}, a state-space sequence model capable of capturing long-range dependencies with linear-time complexity. In contrast to Transformer-based models, which incur $\mathcal{O}(n^2)$ time and memory complexity due to their attention mechanisms, Mamba reduces this overhead to $\mathcal{O}(n)$ by leveraging structured state-space representations. This design yields comparable prediction accuracy while significantly improving training and inference efficiency, making it well-suited for compiler deployment across diverse hardware platforms.
For cross-hardware adaptation, TCL introduces a \textbf{continual knowledge distillation framework} that enables progressive knowledge accumulation from multiple platforms. By decoupling shared knowledge representation (the hardware knowledge base) from hardware-specific adaptation (the hardware active column), the framework avoids the parameter growth and data simultaneity constraints inherent in multi-task learning, enabling more scalable and modular transfer across platforms.
Our key contributions are as follows:

\begin{itemize}
\item We introduce the \textbf{RDU Sampler}, an active learning algorithm that jointly optimizes \textit{representativeness}, \textit{diversity}, and \textit{uncertainty} to select high-value tensor programs. Using only 10\% of the dataset, it attains performance close to or better than models trained on the full dataset, significantly reducing data collection overheads.

\item We design a new cost model architecture based on the \textbf{Mamba Block}, which efficiently captures long-range dependencies in schedule sequences while significantly reducing computational and memory complexity, from $\mathcal{O}(n^2)$ in Transformer-based models to $\mathcal{O}(n)$. This lightweight architecture enables faster training and improved prediction accuracy with fewer parameters.

\item We propose a \textbf{continual knowledge distillation framework} for scalable cross-hardware adaptation. It consists of a shared hardware knowledge base that accumulates information from multiple source platforms, and a hardware active column that transfers this knowledge to new hardware targets. This progressive design avoids the parameter explosion and simultaneous data dependency issues of conventional multi-task learning.

\item We open-source a large-scale dataset of tensor programs, collected over several months on an Intel i7-12700F CPU and an NVIDIA GeForce RTX 3080Ti GPU\footnote{\url{https://github.com/booker0415/Large-Scale-Tensor-Program-Dataset-on-RTX-3080-Ti-and-Intel-i7-12}}. Extensive experiments on this dataset demonstrate that \textbf{TCL} substantially reduces tuning time and improves inference latency compared to state-of-the-art baselines.
\end{itemize}

%% file: 2_Related_works.tex
\section{Related Works}
\textbf{Active learning}~\cite{wu2018pool,shen2023thermal} reduces the annotation workload by selectively labeling the most informative samples, enabling models trained on these subsets to achieve high performance with less data. 
While active learning has been extensively studied in classification settings~\cite{li2024unlabeled, ding2022hyperspectral, shen2022tbal, hemmer2022deal}, its application to regression tasks remains relatively limited~\cite{bemporad2023active, gal2017deep}. 
In the context of DL compilers, offline cost model training typically requires large-scale tensor program datasets. 
Although effective, recollecting such datasets for each new hardware platform incurs significant time and computational costs. To address this challenge, recent studies have investigated active learning and sampling strategies within DL compilers. 
ALT~\cite{zeng2020alt} was the first to integrate active learning into a DL compiler, effectively reducing tuning time in online cost model training. BALTO~\cite{bi2022balto} proposed a biased-diversity selection method to accelerate offline training. CDPMM~\cite{hu2024cdmpp} introduced a clustering-based sampling approach that selects representative tensor programs, improving adaptation to new hardware and enhancing generalization.

\textbf{Cost model architecture} in DL compilers must balance prediction accuracy with computational efficiency. 
Early efforts relied on analytical heuristics; e.g., Halide~\cite{mullapudi2016automatically} estimated performance using data reuse and FLOP counts for loop transformations. 
Subsequent approaches adopted machine learning models: AutoTVM~\cite{chen2018learning} and Ansor~\cite{zheng2020ansor} employed XGBoost to predict relative performance rankings, 
while Tenset~\cite{zheng2021tenset} showed that a combination of MLP and RankLoss yielded superior accuracy. 
More recent works explored deep learning-based architectures. TIRAMISU~\cite{baghdadi2019tiramisu} applied LSTMs to model sequential schedule behavior, whereas One-Shot~\cite{ryu2022one}, TLP~\cite{zhai2023tlp}, and CDPMM~\cite{hu2024cdmpp} adopted Transformer-based encoder-decoder structures. 
Despite their modeling power, LSTMs suffer from limited parallelism, and Transformers incur quadratic complexity with respect to sequence length, making them costly for large-scale tuning scenarios.

\textbf{Cross-hardware} generalization is increasingly critical in DL compiler optimization, as tuning time is dominated by hardware measurements, which often become the primary bottleneck~\cite{qiao2024pruner,shen2022tcps,wasala2025energy}. 
Offline-trained cost models alleviate this by removing the need for on-device profiling, but they are tied to specific platforms and require costly data recollection and retraining for new hardware. 
To address this, transfer learning techniques~\cite{zhang2018overview, long2018transferable, weiss2016survey} have been explored to reuse knowledge from source platforms and accelerate model convergence on target devices. 
Moses~\cite{zhao2023moses}, grounded in the lottery ticket hypothesis, uses domain adaptation to disentangle hardware-aware and hardware-agnostic features for better cross-device generalization. Verma et al.~\cite{verma2023transfer} propose an attention mechanism that embeds transferable hardware-specific signals into cost models. 
TLP~\cite{zhai2023tlp} employs a multi-task learning strategy with shared parameters to learn general optimization patterns and task-specific branches to capture hardware-specific nuances. 
While these methods improve cross-platform tuning efficiency, challenges remain in balancing generalization and scalability.

To bridge these gaps, we propose \textbf{TCL}, a unified framework that integrates: 
(1) the RDU Sampler for data-efficient tensor selection, 
(2) a Mamba-based cost model for fast and accurate prediction, 
and (3) a continual knowledge distillation mechanism that supports scalable and effective cross-platform generalization without multi-tasking overhead.

%% file: 3_Overview.tex
\section{Preliminaries and Overview}
\subsection{Preliminaries}
In this section, we define some key terms and concepts used throughout the paper. These definitions establish a foundation for the proposed methods and analyses.

\textbf{Network and Subgraph.} A DNN is defined with a specific input size, such as batch size and image input dimensions (e.g., ResNet-18, MobileNetV1). A DL compiler partitions the network into subgraphs based on specific rules, where each subgraph represents a part of the original network.

\textbf{Hardware Platform.} The devices on which DNNs are deployed and executed are referred to as hardware platforms. Here, devices with different hardware architectures are considered distinct hardware platforms. We do not transfer knowledge between CPU and GPU in this paper.

\textbf{Assignment.} The execution of a subgraph on a hardware platform is termed an assignment. A DNN can be divided into multiple subgraphs, and when these subgraphs are executed on a hardware platform, they result in multiple assignments. For example, ResNet-18 partitions into 18 subgraphs in Ansor, leading to the creation of 18 assignments.

\textbf{Schedule Primitive.} Schedule primitives are transformations used in DL compilers to optimize the execution of computational graphs across platforms, enhancing performance metrics such as latency, memory usage, and computational efficiency. Commonly used schedule primitives include SP (split), RE (reorder), and FU (fuse), among others.

\textbf{Tensor Program.} A tensor program is a low-level implementation of tensor operations that specifies how computations are performed on a hardware platform.

\begin{figure}[!t]
\centering
\includegraphics[width=\linewidth]{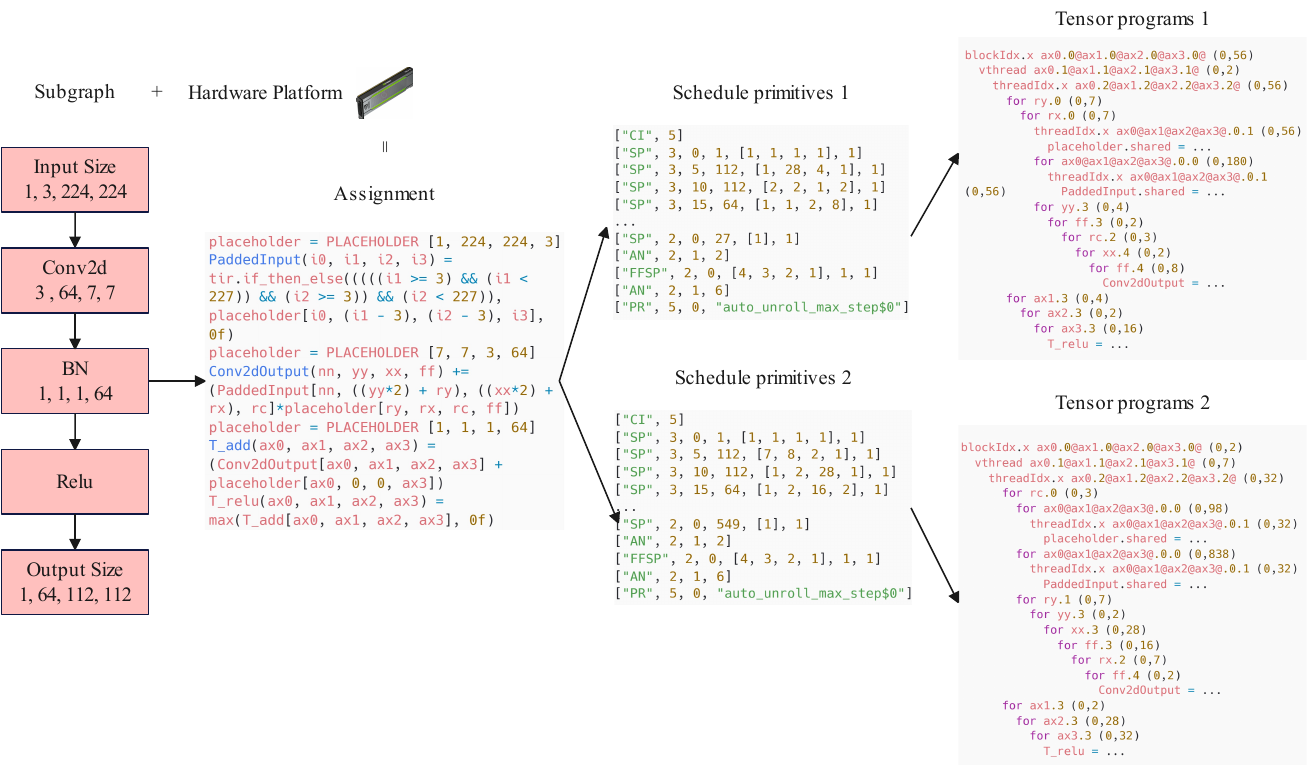}
\caption{Examples of two tensor programs for Conv + BN + ReLU subgraph in ResNet-18 on Nvidia Tesla T4}
\label{fig1}
\end{figure}

We present an assignment for a subgraph (Conv + BN + ReLU) from ResNet-18 on the Nvidia Tesla T4 hardware platform in Fig. \ref{fig1}, along with two variants of the schedule primitives and tensor programs for this assignment.

\subsection{Overview}
\begin{figure}[!t]
\centering
\includegraphics[width=0.8\linewidth]{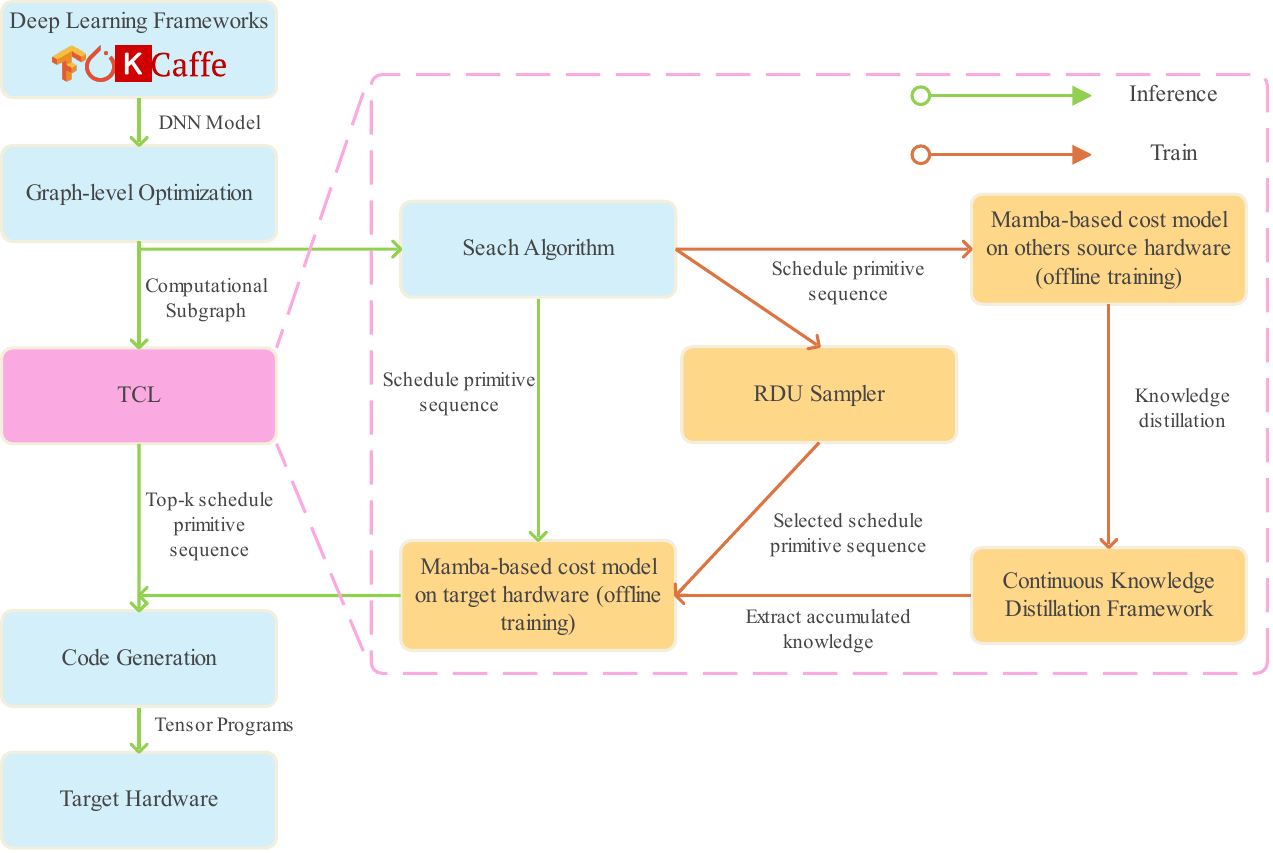}
\caption{Overview of the TCL cross-hardware DL Complier. The training phase offline-trains the cost model for the target hardware, leveraging models from other hardware. The inference phase integrates the trained model into the DL compiler's end-to-end pipeline. The light black blocks represent the modules based on Ansor, the light pink blocks represent our proposed TCL modules, and the light yellow blocks represent the specific components within TCL.}
\label{fig2}
\end{figure}

TCL aims to leverage cost models trained offline on multiple source hardware platforms to facilitate the offline training of cost models on target hardware platforms. 
Fig. \ref{fig2} illustrates the flowchart of TCL. TCL accepts DNN models (e.g., PyTorch\cite{paszke1912pytorch}, TensorFlow\cite{abadi2016tensorflow}, Keras\cite{chollet2015keras}, and Caffe\cite{jia2014caffe}) developed in various DL frameworks. It then performs graph-level optimizations (e.g., operator fusion, constant folding, and data layout transformation) to output a computational subgraph. The granularity of the subgraph is the same as that of the Ansor, and the DNN is converted into a small subgraph using the operator fusion algorithm of Relay\cite{roesch2019relay}. This subgraph is processed by a search algorithm to generate schedule primitive sequences, which are used for training the cost model. Subsequently, TCL operates in two phases: training and inference.

During the training phase, the offline training of the cost model on the target hardware platform relies on a continuous knowledge distillation framework that accumulates knowledge from cost models trained offline on multiple source hardware platforms. The trained cost model on each source device accumulates knowledge into the continuous knowledge distillation framework through knowledge distillation, akin to how the human brain accumulates old knowledge to facilitate the learning of new knowledge. When training the cost model for the target hardware platform, in addition to using the continuous knowledge distillation framework, it is also necessary to use the RDU Sampler for data sampling to save data collection time.

In the inference phase, schedule primitive sequences generated by the search module predicts the performance of the tensor program using the cost model. The cost model selects the best K optimal schedule primitive sequences, generates the tensor program, and deploys it on the target hardware platform. The cost model effectively identifies the relationships between schedule primitive sequences through an encoder-decoder designed based on the Mamba block, focusing on sequences that significantly impact performance.

%% file: 4_Sampling.tex
\section{RDU Sampler}
The RDU Sampler is a key component of TCL, designed to select the most informative and valuable samples from the hardware platform. 
This process reduces data collection time during offline training while ensuring the predictive accuracy of the cost model. 
Since the performance of the cost model relies heavily on the quality of the samples chosen by the RDU Sampler, 
it is essential to carefully design the sampling criteria. 
In our RDU Sampler, we develop a high-performance sampling algorithm based on three criteria: representativeness, diversity, and uncertainty.

\begin{figure}[!t]
\centering
\includegraphics[width=\linewidth]{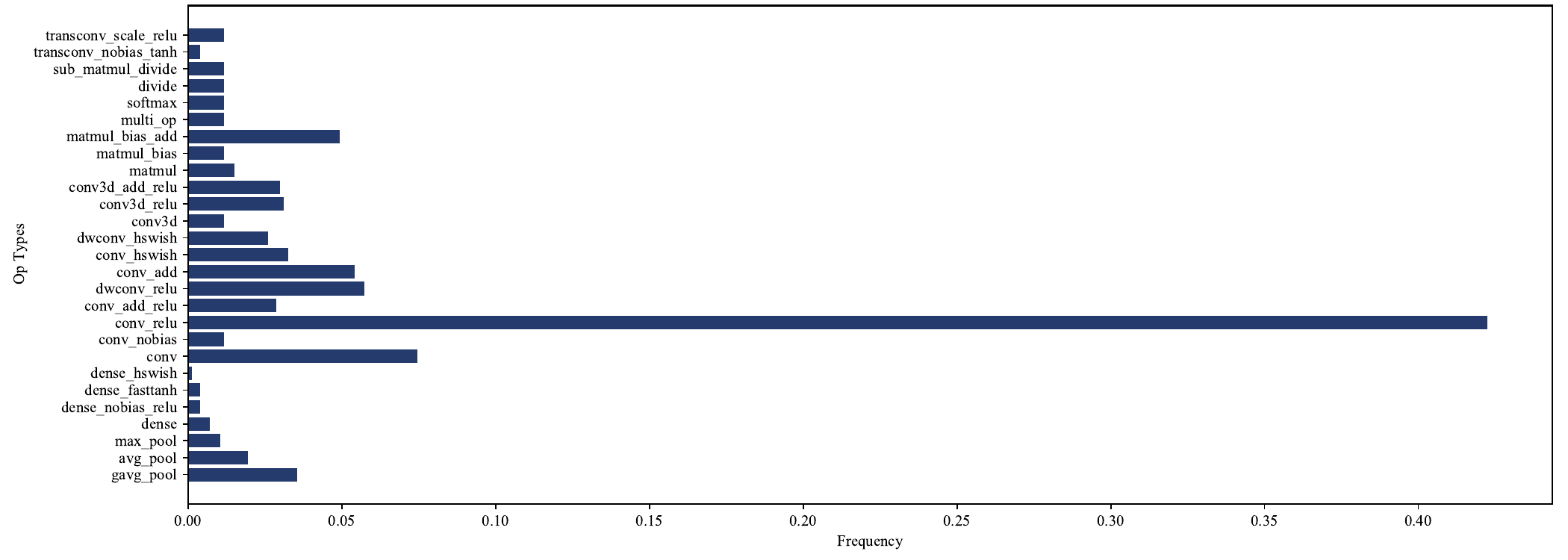}
\caption{Distribution of operator types for assignments on Intel Xeon E5-2673 in the Tenset dataset. This dataset covers 2,308 assignments from multiple deep learning models.}
\label{fig3}
\end{figure}

\textbf{Representativeness} should accurately reflect the characteristics of the entire dataset distribution, ensuring that the sampled data retains a similar class or feature distribution to the original dataset. 
Fig. \ref{fig3} shows the distribution of operator types across assignments in the entire dataset on the Intel Xeon E5-2673 CPU hardware platform. 
It reveals that assignments related to convolution and matrix multiplication constitute the majority, while those related to activation functions such as pooling and softmax account for only a small portion. 
Therefore, during sampling, we aim to ensure that the sampled data preserves this original distribution pattern. We classify assignments based on the operator types within the computational subgraphs and ensure that the class distribution of the sampled assignments aligns with the original dataset distribution.
The $Calc\_Dist$ function in Algorithm \ref{euclid} computes the operator distribution in the original assignments to guide sampling with matching types.

\begin{figure}[!t]
\centering
\includegraphics[width=0.5\linewidth]{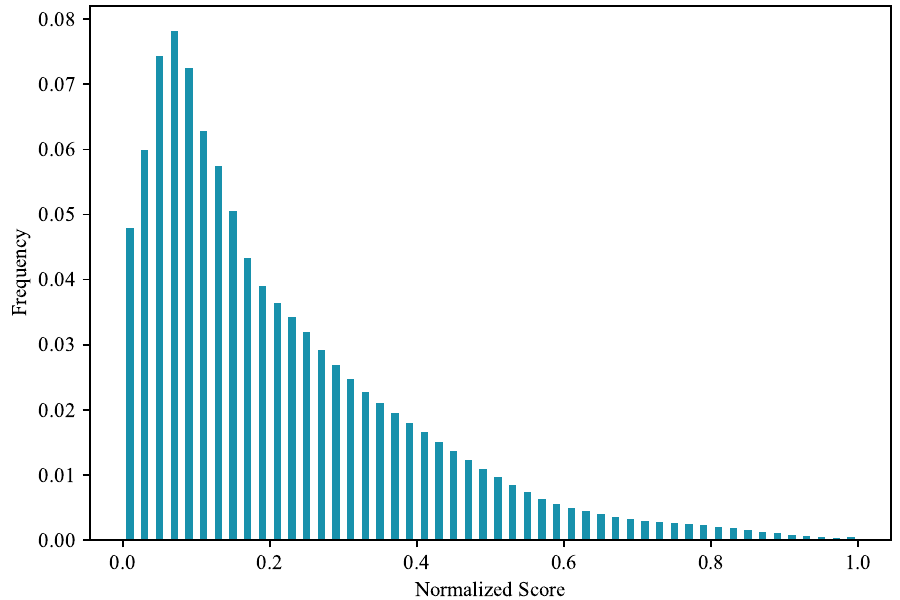}
\caption{Normalized latency score frequency distribution for tensor programs on Intel Xeon E5-2673 in the Tenset dataset. The normalized latency score refers to the ratio between the latency of tensor programs generated using different schedule primitives on the hardware platform and the minimum latency achieved for that task across all schedule primitives.}
\label{fig4}
\end{figure}

\begin{algorithm}[!t]
    \renewcommand{\algorithmicrequire}{\textbf{Input:}}
    \renewcommand{\algorithmicensure}{\textbf{Output:}}
    \caption{RDU sampling algorithm}\label{euclid}
    \begin{algorithmic}[1]
        \Require $f$, uninitialized cost model; $D_u$, unlabeled dataset; $B_t$, total budget of samples at each step $t$. 
        \Ensure $D_l$, labeled dataset. 
        \Function{Calc\_Dist}{$D_u$}
            \State $freq\_map, probs\_map = \{\}$
            \State $total\_assignments = length(D_u)$
            \For{$assignment$ \textbf{in} $D_u$}
                \State $op = get\_op(assignment)$
                \State $freq\_map[op] = freq\_map.get(op,0)+1$ 
            \EndFor
            \For{$op, count$ \textbf{in} $freq\_map$}
            \State $probs\_map[op]$ = $\frac{count}{total\_assignments}$
            \EndFor
            \State \textbf{return} $probs\_map$
        \EndFunction
        \Function{SAMPLE\_MOST\_INFO}{$f$, $D_u$, $B_t$}
        \State $D_l\gets Rand\_Select(D_u)$
        \State $f \gets Init(f)$
        \State $pro\_map = Calc\_Dist(D_u)$
        \For{$op\_type, probability$ \textbf{in} $pro\_map$}
        \State $budget[op\_type] = B_t\times probability$
        \EndFor
        \For{$1$ \textbf{to} $T$}
        \While{$S_t \leq B_t$}
        \State $d\_s(x_i) \gets$  Calculated by Eq. \ref{eq1}
        \State $u\_s(x_i) \gets$ Calculated by Eq. \ref{eq2} and \ref{eq3}
        \State $t\_s(x_i) = \; f(x_i)(d\_s(x_i)) + u\_s(x_i) , x_i \in D_u$
        \State $T\_i = find(\mathop{\operatorname{argmax}_{x_i \in D_u} t\_s(x_i}))$
        \State $T\_op = get\_op(T\_i)$
        
        \If{$selected[T\_op] < budget[T\_op]$}
        \State $D_l\gets D_l\cup\{T\_i\}$
        \Else
        \State continue
        \EndIf
        \EndWhile
        \State $f.update()$
        \EndFor
        \EndFunction
        \end{algorithmic}
\end{algorithm}

\textbf{Diversity} refers to selecting samples that exhibit significant differences from one another to ensure comprehensive coverage of the sample space. 
Fig. \ref{fig4} shows the normalized performance distribution of all samples on the Intel Xeon E5-2673 CPU platform. 
Most samples (91.9$\%$) exhibit a normalized performance below 0.5, while only a small fraction (8.1$\%$) exceed this threshold. The predominance of low-performance tensor programs is unfavorable for training a cost model capable of accurately identifying high-performance ones.
Therefore, to improve the predictive accuracy of the cost model, the selected samples for labeling should balance the distribution of low-performance and high-performance programs. 
The distance between samples is calculated using the formula provided in Eq. (\ref{eq1}), where we compute the score differences between an unlabeled program $x_i$ and each labeled program $x_j$, and take the minimum value as the distance between the unlabeled sample $x_i$ and all labeled samples.

\begin{equation}
    dist(x_i,x_j) = \mathop{\min \left| f(x_i)-f(x_j) \right|} \limits_{x_i \in D_u,x_j \in D_l}
\label{eq1}
\end{equation}

Here, $j=1, \ldots, M, i= M+1, \ldots, N$, $D_u$ is the unlabeled dataset, $D_l$ is the labeled dataset. $f$ is a cost model used to predict the performance of tensor programs. $f(x_i)$ and $f(x_j)$ are both predictions from the cost model and are normalized before being used in the computation.
We select samples with larger $dist(x_i,x_j)$ to promote diversity. 
When distances are equal, priority is given to programs with higher predicted performance, as the cost model aims to identify high-performance candidates.

\textbf{Uncertainty} mainly helps the model to select samples for labeling that it currently predicts with the greatest uncertainty and carries the most information. 
These samples are typically those that the model finds difficult to predict accurately given the existing training data, thereby facilitating rapid improvement and enhanced performance of the model. 
Common metrics for quantifying uncertainty include prediction confidence, entropy, prediction loss, and variance. 
Since our sampling task is a regression problem, we use the variance of the predicted values between the unlabeled sample $x_i$ and each labeled sample $x_j$ as the uncertainty metric. 
The calculation formulas are provided in Eq. (\ref{eq2}) and Eq. (\ref{eq3}).


\begin{equation}
    \mu = \frac{f(x_i) + \sum_{j=1}^{M} f(x_j)}{M + 1}, \quad x_i \in D_u, x_j \in D_l
\label{eq2}
\end{equation}

\begin{equation}
    \sigma^{2} = \frac{(f(x_i) - \mu)^2 + \sum_{j=1}^{M} (f(x_j) - \mu)^2}{M + 1}
\label{eq3}
\end{equation}

Algorithm \ref{euclid} illustrates the detailed steps of our RDU sampling algorithm. 
Lines 1-12 define a function that calculates the distribution of operator types across all assignments in the original unlabeled dataset. 
Line 2 initializes $freq\_map$ to count the occurrences of each operator type and $probs\_map$ to calculate their probabilities. 
Line 3 calculates the total number of assignments in the unlabeled data pool. Lines 4-7 count the frequency of each operator type across assignments in the unlabeled data pool, and lines 8-12 calculate the probability of each operator type in the pool. 
Lines 13-35 define a sampling function to select the most valuable samples from the unlabeled data pool. 
Line 14 randomly initializes a batch of samples, line 15 initializes the cost model using the samples from line 14, and line 16 calls the $CALC\_DIST$ function to calculate the operator type distribution of all unlabeled samples. Lines 17-19 calculate the budget for selecting operator types over $t$ iterations. 
Lines 20-23 compute the scores for the distance and variance according to Eq. \ref{eq1}-\ref{eq3}. 
In line 24, we give a weight $f(x_i)$ to $d\_s$ which is used to preferentially select samples with higher scores when $d\_s$ is the same, as our cost model's objective is to identify high-quality tensor programs. 
The sample with the highest overall score is then used for training. Lines 25-31 assess whether the operator types of the selected sample have reached their budget limits. 
The sample is selected if the limits are not reached; otherwise, it is excluded. Finally, line 33 updates the cost model.

%% file: 5_MambaCost.tex
\section{Mamba-based Cost Model}

\subsection{Feature Extraction}

Carefully designed feature engineering is the cornerstone to the success of cost models. Ansor\cite{zheng2020ansor} extracts 164 features from the innermost assignment statements, focusing on five key aspects: computation, memory access, arithmetic intensity, and more. Tenset-MLP\cite{zheng2021tenset} adds 10 high-level computation graph intermediate representation (IR) features to Ansor's feature extraction. 
TIRAMISU\cite{baghdadi2019tiramisu} uses abstract syntax tree (AST) structures as input features. TLP\cite{zhai2023tlp} is the first to propose feature extraction on schedule primitives and represents a SOTA feature extraction method. 
Our feature representation method is an improvement based on TLP. TLP extracts schedule primitive features from tensor programs (hardware-independent), which remain consistent across different hardware platforms, thereby supporting multi-task learning and leveraging knowledge from multiple source devices. However, for our continual knowledge distillation framework, purely hardware-independent features are insufficient for efficient transfer. To address this, we enhance the TLP-based feature design by introducing hardware-aware features, which are specifically tailored for each hardware platform. These features enable the cost model to learn the relationships between schedule decisions and execution outcomes across different platforms. As shown in Table~\ref{table1}, we extract computation- and memory-related features separately for CPU and GPU, and this approach can be easily extended to other hardware platforms. Consequently, each schedule primitive sequence has a feature dimension of 26$\times$22 on CPU and 41$\times$20 on GPU.


\begin{table}[!t]
    \centering
    \caption{CPU and GPU Feature Parameters}{}
{
    \begin{tabular}{cc}
        \toprule
        CPU Features & GPU Features \\ 
        \midrule
        Cores & Cuda Cores \\
        Threads & Base Clock \\
        Turbo frequency & Boost Clock \\
        Boost Frequency & Memory clock \\
        L1 Cache & Memory size\\
        L2 Cache & Memory bus width \\
        L3 Cache & Peak Memory bandwidth \\
        Max memory channels & FP64 Performance \\
        Max memory capacity & FP32 Performance \\
        Max memory bandwidth & FP16 Performance \\
        Bytes Size of SIMD Extensions & INT8 Performance \\
       - & INT4 Performance \\
        \bottomrule
    \end{tabular}
    \label{table1}
}
\end{table}

\begin{figure}[!t]
\centering
\includegraphics[width=0.45\linewidth]{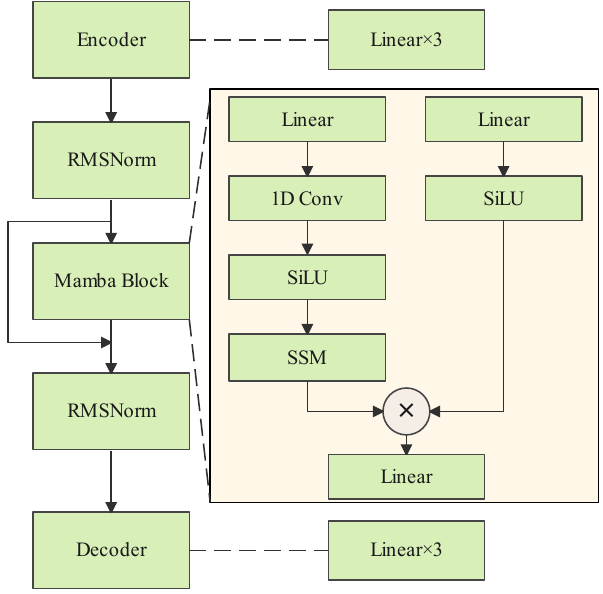}
\caption{Model architecture description of Mamba-based cost model.}
\label{fig5}
\end{figure}

\subsection{Model Architecture}
The current SOTA cost model is TLP, which is based on the Transformer architecture and has achieved excellent performance in DL compilers. However, the Transformer architecture introduces significant computational and memory overhead. 
Mamba \cite{gu2023mamba} excels at sequence processing, effectively capturing long-range dependencies while significantly reducing computational and memory overhead, thus shortening the runtime of the cost model. Moreover, its input-dependent dynamic filtering mechanism enables selective updates of relevant states or attention components based on the input. This facilitates capturing schedule primitives that strongly affect inference latency.
Mamba is based on a structured state-space model (SSM\cite{gu2023modeling}). SSM is a continuous system that takes a continuous input sequence $\mu(t)$ and maps it to a continuous output sequence $y(t)$ through a latent state $z(t)$ The mathematical definition of the system is as follows:

\begin{equation}
    z'(t) = A z(t) + B \mu(t)
\label{eq4}
\end{equation}

\begin{equation}
    y(t) = C z'(t)
\label{eq5}
\end{equation}

In Eq. (\ref{eq4}), $A \in \mathbb{R}^{N \times N}$ represents the state transition matrix, which captures the changes in state over time. $B \in \mathbb{R}^{N}$ represents the weight matrix that maps the input to the state, while $C \in \mathbb{R}^{N}$ represents the observation matrix mapping the latent hidden state to the output. 
S4\cite{gu2021efficiently} applies the above continuous system to discrete sequences using a discretization method. It accelerates inference with a recurrent structure, speeds up training with a convolutional structure, and addresses long-range dependencies in sequence modeling using the HiPPO\cite{gu2020hippo} matrix. 
Mamba enhances S4 by utilizing a selection mechanism that efficiently identifies relevant input information based on the input itself, along with implementing a hardware-aware algorithm to improve the model's computational efficiency and runtime performance on hardware platforms.

Fig. \ref{fig5} illustrates our cost model architecture based on Mamba. 
First, the input is mapped to high-dimensional features through an encoder composed of three linear layers. 
After normalization, the encoder output is fed into the Mamba Block, which captures dependencies between sequences while maintaining output dimensions consistent with the input. 
Finally, the output is normalized again and passed to the decoder to obtain the predicted results. The encoder comprises three linear layers with output dimensions of 64, 128, and 128, while the decoder consists of three linear layers with output dimensions of 64, 32, and 1. The hyperparameters of the Mamba architecture in our cost model are determined through experiments described in section~\ref{sec：Hyper Mamba}. 

\subsection{Loss Function}
Two types of loss functions are commonly used in cost models for DL compilers: Mean Squared Error (MSE) and Ranking loss. 
With MSE as the loss function, the trained cost model tends to fit the actual latency of tensor programs on specific hardware platforms. 
However, our goal is to identify high-performance tensor programs rather than to accurately predict the actual latency of tensor programs, so MSE loss is not suitable for training the cost model.
Therefore, we employ LambdaRank\cite{wang2018lambdaloss} loss to focus on high-performance tensor programs, aiming to find the tensor programs with the lowest inference latency on hardware platforms. 
The LambdaRank loss function is defined as follows:

\begin{equation}
    L_{Rank}=-\sum_{i,j}\Delta\mathrm{NDCG}(i, j)\log_2(1 + e^{-\sigma (f(x_i) - f(x_j))})
\label{eq6}
\end{equation}

where $\Delta \mathrm{NDCG}(i, j) = |G_i - G_j| \cdot \left| \frac{1}{D_i} - \frac{1}{D_j} \right|$ guides the direction for improving ranking performance. $G_i = \frac{2^{y_i} - 1}{\text{maxDCG}}$ is a gain function, which aims to prioritize high-performance programs in the ranking. $D_i = \log_2(1 + i)$ is a discount function that reduces the impact of lower-ranked programs on the model. 
In $G_i$, $\text{maxDCG}$ represents the maximum gain of a ranking in an ideal scenario, derived from the gain calculated using the real latency of the programs.

%% file: 6_Knowledge.tex
\section{Continuous Knowledge Distillation Framework}

\begin{figure}[!t]
\centering
\includegraphics[width=0.8\linewidth]{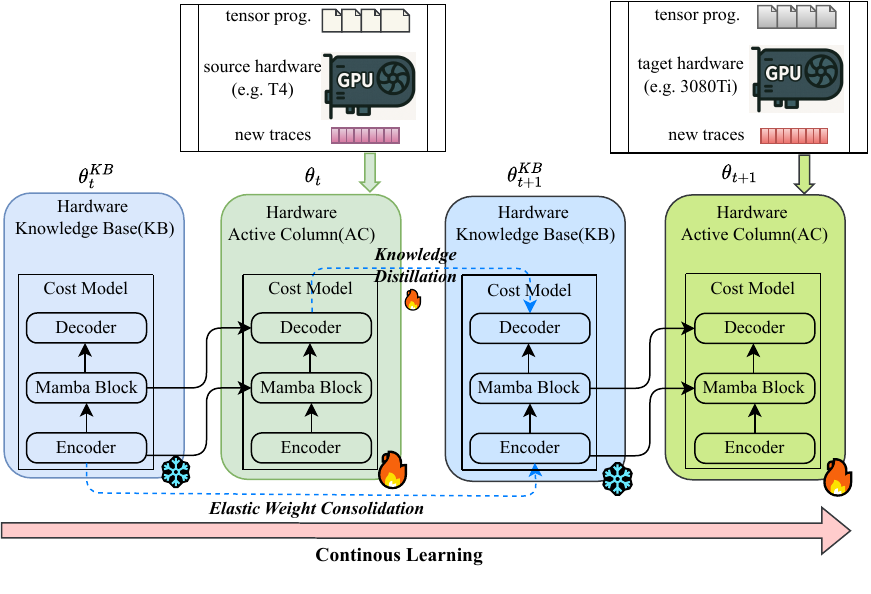}
\caption{Overview of the proposed Continuous Knowledge Distillation (CKD) framework for cross-hardware cost model training. For each new hardware platform (e.g., CPU, GPU), an \textbf{active column (AC)} learns platform-specific knowledge from new tensor program traces while the \textbf{knowledge base (KB)} remains frozen to preserve prior knowledge. After the continual learning (CL) phase, knowledge from the AC is selectively distilled into the KB via a gated distillation mechanism and constrained by Elastic Weight Consolidation (EWC) to prevent catastrophic forgetting. This process enables progressive, scalable adaptation across diverse hardware targets without requiring simultaneous multi-platform data or expanding model size. Both the AC and KB share the same Mamba-based architecture (encoder, Mamba block, decoder).}
\label{fig6}
\end{figure}

An increasing number of researchers are focusing on the offline training of cost models across hardware platforms. 
Most existing methods are based on transfer learning, where knowledge from cost models trained offline on a single hardware platform is transferred to others. 
MTL-TLP\cite{zhai2023tlp} is the first method to use multi-task learning to leverage knowledge from multiple hardware platforms, thereby improving the offline training of cost models for a target platform. 
However, as the number of hardware platforms increases, both the number of parameters (Params) and the floating point operations (FLOPs) required by the model increase significantly, while also requiring input data from multiple hardware platforms to train the cost model. 
To tackle the challenges of cross-hardware platforms, we propose a framework based on continuous knowledge distillation (CKD), as shown in Fig. ~\ref{fig6}. This framework efficiently leverages knowledge from multiple source hardware platforms without increasing model size or computational cost, and it does not require, as in multi-task learning, the simultaneous input of large amounts of data, thereby avoiding additional storage and computational overhead. The cross-hardware generalization capability is achieved through continual learning (CL) and knowledge distillation (KD): the cost model trained on the source device stores hardware-independent knowledge (such as the relationship between scheduling primitives and inference performance) into a knowledge base (KB) via KD, and during the CL stage, the knowledge in the KB is leveraged to facilitate learning on the target hardware.
Our approach features a progressive network architecture composed of a hardware KB and hardware active column (AC), alternating between two training phases: CL and KD. 
During the CL phase, the hardware KB is frozen to prevent forgetting previously acquired knowledge from other hardware platforms, while the hardware AC is used to learn knowledge about the new hardware platform and leverage the knowledge from the hardware knowledge base through their connections.
During the KD phase, the hardware AC is frozen to distill its knowledge into the hardware KB. At the same time, Elastic Weight Consolidation (EWC)\cite{kirkpatrick2017overcoming} is applied to prevent forgetting the knowledge already stored in the hardware KB.
In our framework, the hardware KB and the hardware AC form an integrated whole, both implemented using the Mamba-based cost model shown in Fig.~\ref{fig5}. KB and AC share the same architecture, resulting in a lightweight model with a total size of 0.7 MB. During the alternating phases of CL and KD, their parameters are iteratively updated.

When training the cost model on a new hardware platform, the hardware KB in the CKD framework remains frozen.
The hardware AC integrates the outputs from each layer of the hardware KB network with the corresponding outputs from each layer of the AC network through a progressive network, feeding them together into the next layer of the AC network to reuse the old knowledge. 
$h_i$ represents the activation output layer of a Mamba block, encoder, or decoder in the cost model. The superscript $^{\textbf{KB}}$ refers to the KB, and $\sigma$ is a nonlinear activation function. The output of the $i$-th layer in the AC is as follows:

\begin{equation}
    h_i = \sigma\left(W_i h_{i-1} + \alpha_i \odot U_i \sigma\left(V_i h_{i-1}^{KB} + c_i\right) + b_i\right)
\label{eq7}
\end{equation}

where, $b_i$ and $c_i$ are the biases, $\alpha_i$ is a trainable vector with the same size as the number of units in the $i$-th layer, and $W_i$, $U_i$,$V_i$ are the weight matrices. 
The symbol $\odot$ represents element-wise multiplication. 
In the CL phase, we focus on learning a cost model for a new hardware platform, and the positive transfer of old knowledge is achieved through network connections. 
Therefore, the loss function for this phase is defined as follows:

\begin{equation}
L_{\text{CL}} = L_{\text{Rank}}(p_t, l_t; \theta_t)
\label{eq8}
\end{equation}


Here, $p_t$ is the prediction of the cost model on the $t$-th hardware platform, $l_t$ is the latency label on the $t$-th hardware platform, and $\theta$ is  the parameter of the hardware AC when learning the cost model for the $t$-th platform.

When extracting the knowledge of a cost model trained on a new hardware platform into the hardware KB, the CKD framework uses KD to transfer the knowledge from the hardware AC to the hardware KB. 
In this process, the hardware AC acts as the teacher model, while the KB serves as the student model. 
To prevent the KB from forgetting previously stored knowledge, we introduce a forgetting prevention mechanism. 
This mechanism is implemented through the EWC algorithm, which constrains the changes to the KB parameters during distillation, ensuring that the updates remain within the previous solution space, thereby effectively preserving the existing knowledge. 
This ensures that the KB continues to accumulate new knowledge while retaining the memory of previously learned information. 
The teacher model generally demonstrates high predictive accuracy and possesses strong generalization capabilities in most cases. 
However, in practice, it may also yield numerous misleading predictions.
Therefore, it is imperative to impose restrictions on the knowledge transferred by the teacher model to ensure that the student model retains its capacity for independent learning. Consequently, the loss function designed for the distillation phase is expressed as follows:

\begin{equation}
L_{\text{KD}} = \sum_{j} \beta \cdot L_{\text{Rank}}(P_{Sj}, P_{Gj}; \theta_t^{\text{KB}}) + \sum_{j} (1 - \beta) \cdot \Phi_j \cdot L_{\text{Rank}}(P_{Sj}, P_{Tj}; \theta_t^{\text{KB}})+  \sum_{k} \frac{\lambda}{2} F_{t,k} \left( \theta_{t,k}^{\text{KB}} - \theta_{t-1,k}^{\text{KB}} \right)^2 
\label{eq9}
\end{equation}

\begin{equation}
\Phi_j = 1 - \frac{ L_\text{Rank}(P_{Tj}, P_{Gj}; \theta_t^{\text{KB}}) - \min(e_T)}{ \max(e_T) - \min(e_T) }
\label{eq10}
\end{equation}

\begin{equation}
e_T = \{ L_\text{Rank}(P_{Tj}, P_{Gj}; \theta_t^{\text{KB}}) : j = 1, \ldots, N \}
\label{eq11}
\end{equation}

The $L_{\text{KD}}$ loss consists of three components: 
the first measures the student's independent learning ability,
the second assesses the loss of knowledge learned by the student from the teacher,
and the third determines the degree of forgetting of previously acquired knowledge. 
Here, $\Phi_j$ represents the normalized score of the teacher loss for the $j$-th batch of samples, while $e_T$ denotes the teacher loss for each batch across the entire dataset.
$\beta$ is the coefficient that balances the student loss and the distillation loss, and $\theta_t^{\text{KB}}$ refers to the parameters of the KB for the $t$-th task. $P_{Sj}$ is the predicted output of the student model for the $j$-th batch, $P_{Tj}$ is the predicted output of the teacher model for the $j$-th batch, and $P_{Gj}$ is the ground-truth label for the $j$-th batch. 
$\lambda$ is the weight factor that regulates forgetting, and $F_{t,k}$ is the Fisher matrix corresponding to the $k$-th parameter in the KB after storing the knowledge of the $t$-th hardware platform. In our experiments, we set $\beta$ to 0.5 and $\lambda$ to 10,000.

%% file: 7_Experimental.tex
\section{Experiments and Results}
\subsection{Experimental Setup}
\subsubsection{Dataset}
Our dataset is constructed from both the open-source Tenset dataset collected from six different hardware platforms (four CPUs and two GPUs) and additional data collected on our own hardware platforms, specifically Intel i7-12700F CPU and Nvidia GeForce RTX 3080Ti GPU. \textcolor{black}{Data collection follows the Tenset standard protocol, and all results are obtained by averaging multiple measurements to improve stability and reliability.} Each CPU platform contains about 8.65 million samples, and each GPU platform contains about 8.44 million samples. Accordingly, 10$\%$ of the data corresponds to approximately 0.86 million and 0.84 million samples, respectively. The complete dataset covers a wide range of neural network architectures with varying input sizes, including ResNet-18, ResNet-50, ResNeXt-50, Wide-ResNet50, MobileNetV2, MobileNetV3, InceptionV3, DenseNet-121, ResNet3D-18, VGG-16, BERT, and DCGAN.  For each platform-specific dataset, we designate a test set consisting of five representative neural network models: ResNet-50, MobileNet-V2, ResNeXt-50, BERT-tiny, and BERT-base, with a batch size of 1 and an input image size of 224 (or sequence length of 128 for BERT models). The remaining samples are split into training and validation sets with a ratio of 9:1.
\subsubsection{Evaluation Metrics}
We use two evaluation metrics: dataset-based evaluation and end-to-end evaluation.
\textbf{Dataset-based evaluation} adopts the Top-k score, as defined in Eq. (\ref{eq12}), to assess the cost model’s ability to predict the latency ranking of tensor programs. A higher Top-k score indicates a closer match between the predicted and actual latency rankings, reflecting better prediction accuracy. \textcolor{black}{The Top-k score is a key metric for evaluating the performance of cost models in deep learning compilers. Tenset demonstrates that, compared to RMSE and Pairwise Accuracy, it more accurately reflects the predictive ability for high-performance tensor programs. This metric has been adopted in TLP, and we also employ it in our work.}

\begin{equation}
    Top-k=\frac{\sum_{m,s} min\_latency_{m,s} \cdot weight_{m,s}}{\sum_{m,s} \min_{i \in [1,k]} p\_latency_{m,s,i} \cdot weight_{m,s}}
\label{eq12}    
\end{equation}

$min\_latency_{m,s}$ is the lowest latency across all tensor programs corresponding to subgraph $s$ within model $m$. $weight_{m,s}$ is the occurrence frequency of subgraph $s$ in model $m$. Meanwhile, $p\_latency_{m,s,i}$ is the latency of the i-th candidate tensor program for subgraph $s$ in model $m$, ranked according to the cost model’s predicted scores.


\textbf{End-to-end evaluation} reflects the efficiency and performance of DL compilers in optimizing DNNs. An efficient DL compiler should achieve comparable inference latency within a shorter tuning time, and deliver better latency optimization results within the same tuning time.

\subsubsection{Experiment Environment}
The dataset-based evaluation is performed on a server equipped with an Intel Xeon Silver 4214 CPU with 256 GB memory, and four NVIDIA 32 GB V100 GPUs. The training environment for the cost model was composed of Python 3.8.0, PyTorch 1.7.0, CUDA 10.1.243, and cuDNN 7.6.0. The cost model was trained for 100 epochs using the Adam optimizer, with an initial learning rate of 0.0007 and a batch size of 1024, \textcolor{black}{and no early stopping was applied.} For the CKD framework, the epoch scheduling strategy is evenly divided according to the hardware platform and the number of CL and KD stages. The learning rate follows a cosine annealing schedule with a period of 10 during the CL stage, and a stepwise decay strategy during the KD stage, decreasing by 0.1 every 5 epochs. The end-to-end evaluation was carried out on two distinct hardware platforms. One is a CPU platform with an Intel i7-12700F processor, and another is a GPU platform equipped with an NVIDIA GeForce RTX 3080 Ti. TCL was integrated into TVM (version 0.8.0) and evaluated under an environment consisting of Python 3.8.20, PyTorch 1.8.0, CUDA 11.1, and cuDNN 8.2.0.

\subsection{Dataset-based Evaluation of Mamba-based Cost Model}
\subsubsection{Experiments of Hyperparameters in the Mamba Architecture}
\label{sec：Hyper Mamba}
The Mamba architecture leverages a SSM to unfold feature maps along the spatial dimension into sequences, enabling efficient global information integration. In our proposed cost model, the Mamba block includes several key hyperparameters. These are the state expansion factor $d_{state}$, the convolution width $d_{conv}$, the block expansion factor $expand$, and the number of Mamba Blocks $n_{layers}$. The state expansion factor $d_{state}$ defines the dimensionality of the internal state space used by the state space model. The convolution width $d_{conv}$ determines the kernel size of a lightweight depthwise one-dimensional convolution layer that precedes the state space model. The block expansion factor $expand$ defines how much the feature dimensions are expanded within each Mamba block. The number of Mamba blocks $n_{layers}$ specifies how many such blocks are stacked in the overall architecture. To determine the optimal combination of hyperparameters in the cost model, we conducted experiments on an Intel Xeon E5-2673 CPU and a NVIDIA T4 GPU. Table \ref{tab2} presents the settings and results.

As shown in Table \ref{tab2}, we observe that stacking more Mamba blocks, as well as increasing the block expansion factor and state expansion factor, leads to a decline in the prediction accuracy of the cost model. Only increasing the convolution width consistently improves the model's predictive performance. This may be attributed to the fact that an excessive number of Mamba blocks can cause instability in gradient propagation, making convergence difficult—for instance, on the Nvidia T4 GPU. Moreover, larger state and block expansion factors significantly increase the internal representation dimensionality and complexity of the SSM, causing drastic changes in the feature space and making it difficult for the cost model to capture meaningful relationships. In contrast, a wider convolution kernel provides a stronger local receptive field, which helps the cost model perceive structural changes more effectively and capture more stable performance features. The results on a Intel Xeon E5-2673 CPU and a NVIDIA T4 GPU show that the Mamba-based cost model hyperparameter combination of [1, 8, 1, 4] can achieve the best accuracy.

\begin{table}[!t]
\caption{Experiments on different hyperparameter combinations of Mamba-based cost model}
\resizebox{0.79\linewidth}{!}{
\begin{tabular}{cccc|cc|cc}
\hline
          &           &         &         & \multicolumn{2}{c|}{T4} & \multicolumn{2}{c}{E5-2673} \\ \hline
$n_{layer}$  & $d_{state}$  & expand  & $d_{conv}$ & Top-1 Score     & Top-5 Score    & Top-1 Score       & Top-5 Score       \\ \hline
1         & 32        & 1       & 4       & 0.856      & 0.91      & 0.8913       & 0.9605       \\
1         & 16        & 1       & 4       & 0.8848     & \textbf{0.954}     & 0.8530       & 0.9502       \\
1         & 8         & 1       & 4       & \textbf{0.8891}     & 0.9532    & \textbf{0.9050}       & \textbf{0.9639}       \\
2         & 8         & 1       & 4       & 0.8512     & 0.9314    & N/A          & N/A          \\
3         & 8         & 1       & 4       & 0.8458     & 0.9061    & N/A          & N/A          \\
1         & 8         & 2       & 4       & 0.874      & 0.9206    & 0.8903       & 0.9427       \\
1         & 8         & 3       & 4       & 0.8654     & 0.9318    & 0.8839       & 0.9413       \\
1         & 8         & 1       & 3       & 0.8794     & 0.9216    & 0.8828       & 0.9606       \\
1         & 8         & 1       & 2       & 0.8642     & 0.9135    & 0.8696       & 0.9419       \\ \hline
\multicolumn{8}{c}{Table Note: N/A   means did not converge due to instability in optimization.} \\ \hline
\end{tabular}
\label{tab2}
}
\end{table}

\subsubsection{Comparison with Other Architectures}
With the continuous development of DL technology, an increasing number of DL architectures are being applied to predict the performance of tensor programs. Tenset uses an MLP architecture to design a cost model for predicting tensor program performance; TLP is based on the Transformer architecture, while the cost model we designed is based on the Mamba architecture. We evaluated our architecture against the two typical architectures on five CPU platforms and three GPU platforms, and the results are shown in Table \ref{tab3}. 

The results show that TLP achieves the best performance on the Intel Platinum 8272CL CPU platform. On the K80 GPU platform, Tenset achieves the best Top-1 score, while TLP performs best in terms of the Top-5 score. Notably, our Mamba-based cost model achieves the best performance on six out of the eight hardware platforms, while utilizing fewer parameters compared to the other models. This indicates that our Mamba-based cost model performs better at capturing schedule primitive sequences that are highly correlated with inference latency, which may be attributed to Mamba’s dynamic filtering mechanism that selectively adjusts state transition parameters based on different inputs. Furthermore, our model achieves higher prediction accuracy than Tenset and TLP with a model size of only 0.35 MB.

\begin{table}[!t]
\centering
\caption{Comparison of Cost Models on Different Hardware Platforms}
\resizebox{\linewidth}{!}{%
\begin{tabular}{l|ccc|ccc|ccc}
\hline
\multirow{2}{*}{\makecell[c]{\textbf{Hardware platforms}}} & \multicolumn{3}{c|}{Tenset} & \multicolumn{3}{c|}{TLP} & \multicolumn{3}{c}{Ours} \\
\cline{2-10}
& Top-1 Score& Top-5 Score& Size & Top-1 Score & Top-5 Score& Size & Top-1 Score& Top-5 Score & Size \\
\hline
Intel Platinum 8272CL & 0.8748 & 0.9527 & \multirow{8}{*}{\makecell[c]0.92MB} & \textbf{0.9194} & \textbf{0.9710} & \multirow{8}{*}{\makecell[c]1.32MB} & 0.9132 & 0.9592 & \multirow{8}{*}{\makecell[c]{\textbf{0.35MB}}} \\
Intel E5-2673 v4      & 0.8332 & 0.8977 &         & 0.8941 & 0.9633 &         & \textbf{0.9050} & \textbf{0.9639} & \\
AMD EPYC 7452         & 0.8510 & 0.9175 &         & 0.9055 & 0.9494 &         & \textbf{0.9058} & \textbf{0.9619} & \\
ARM Graviton2         & 0.7799 & 0.9049 &         & 0.8207 & 0.9226 &         & \textbf{0.8300} & \textbf{0.9264} & \\
Intel i7-12700F             & 0.8807 & 0.9475 &         & 0.9068 & 0.9537 &         & \textbf{0.9229} & \textbf{0.9691} & \\
NVIDIA Tesla K80      & \textbf{0.9083} & 0.9629 &         & 0.9059 & \textbf{0.9741} &         & 0.8839 & 0.9661 & \\
NVIDIA Tesla T4       & 0.8757 & 0.9528 &         & 0.8847 & 0.9250 &         & \textbf{0.8891} & \textbf{0.9532} & \\
NVIDIA RTX 3080Ti     & 0.8195 & 0.8744 &         & 0.8346 & 0.8955 &         & \textbf{0.8521} & \textbf{0.9128} & \\
\hline
\end{tabular}%
}
\label{tab3}
\end{table}

\begin{figure}[!t]
  \centering
  \begin{subfigure}[b]{0.49\textwidth}
    \centering
    \includegraphics[width=\textwidth]{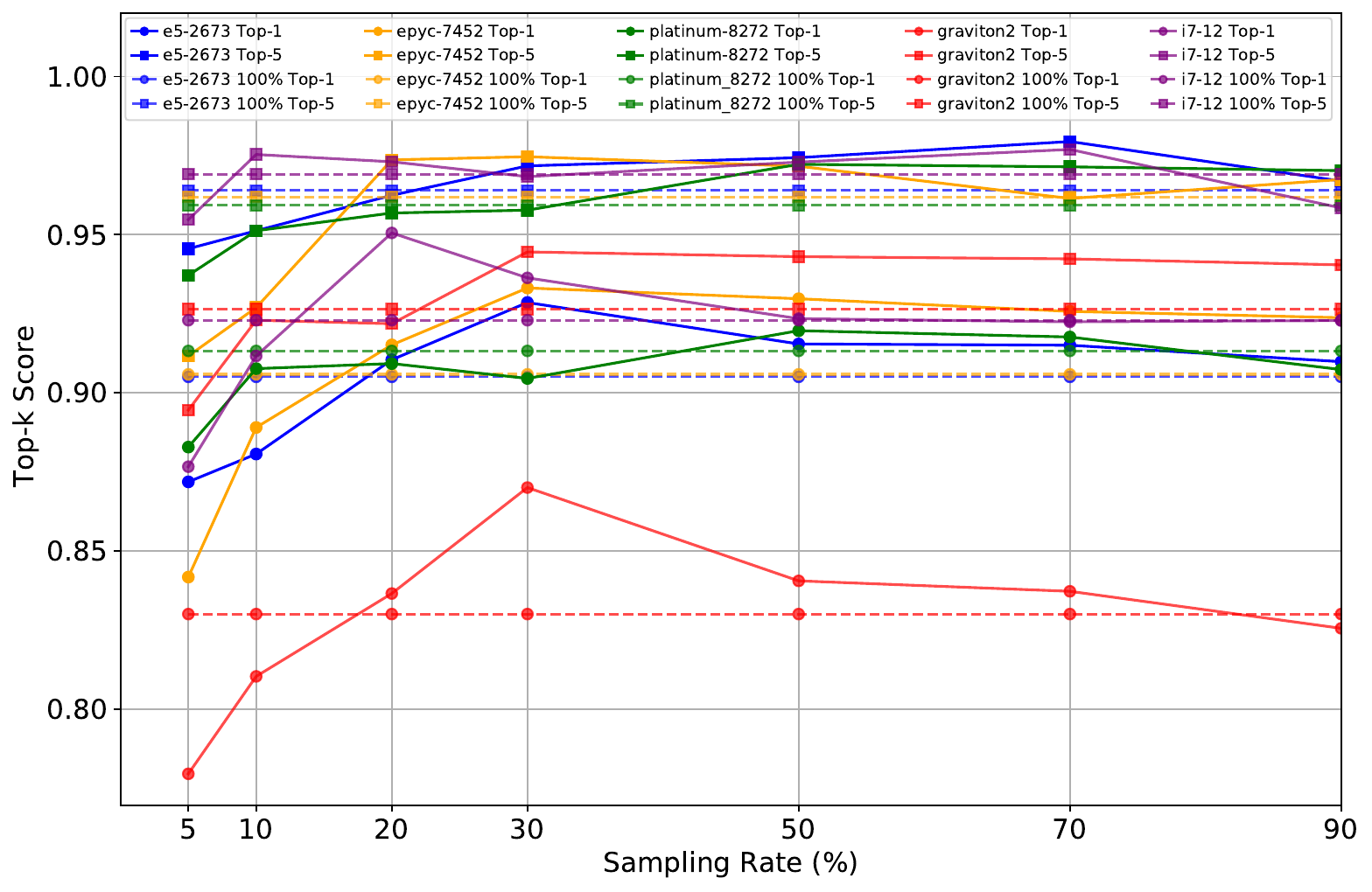}
    \caption{CPU Platforms.}
    \label{fig:a}
  \end{subfigure}
  \begin{subfigure}[b]{0.49\textwidth}
    \centering
    \includegraphics[width=\textwidth]{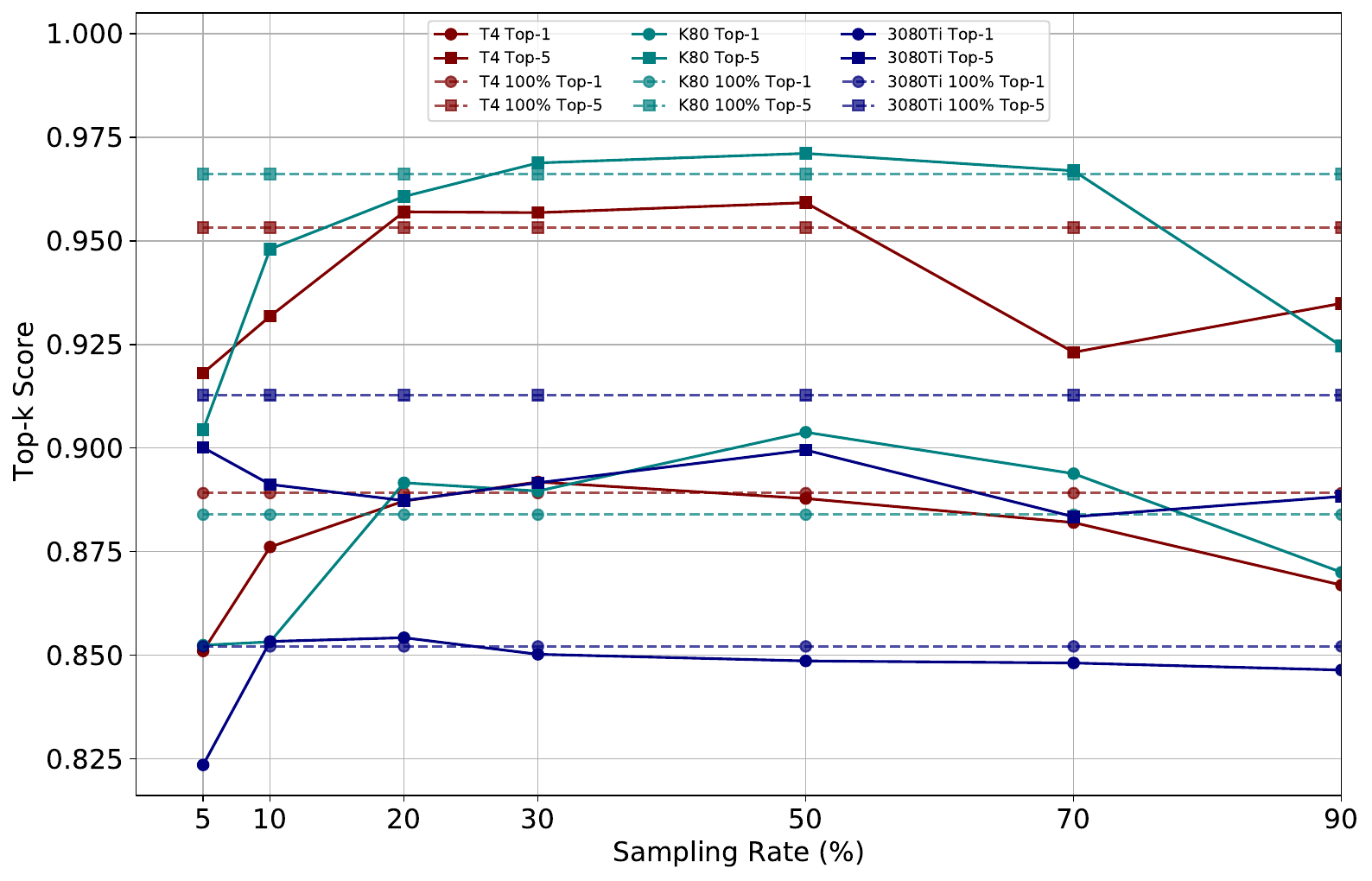}
    \caption{GPU Platforms.}
    \label{fig:b}
  \end{subfigure}
  \caption{Comparison Across Different Sampling Rates on CPU and GPU Platforms.}
  \label{fig7}
\end{figure}

\subsection{Dataset-based Evaluation of RDU Sampler}
\subsubsection{Evaluation of RDU Sampler under Different Sampling Rates}
In general, more data helps improve the performance of DNN models. However, the introduction of excessive low-quality data may negatively affect the model's performance and hinder the training of an efficient model. Therefore, our goal is to train the optimal cost model on the target hardware platform with the lowest possible sampling rate and the shortest data collection time.

To achieve this, we evaluated the performance of the RDU sampler under different sampling rates based on the cost model built with the proposed Mamba architecture, as shown in Fig. \ref{fig7}. The results show that, whether on CPU or GPU hardware platforms, the cost model trained with a 5$\%$ sampling rate exhibits a significant gap in Top-1 and Top-5 scores compared to the model trained on the full dataset (100$\%$). In contrast, the model trained with a 10$\%$ sampling rate achieves Top-1 and Top-5 scores close to those of the full-data model on most platforms, and even surpasses it on some platforms. Compared with the 5$\%$ sampling rate, it demonstrates higher sampling efficiency and better model performance. Therefore, in TCL, when transferring to the target hardware platform and collecting data, we only use a 10$\%$ sampling rate to collect data on target hardware platform, balancing sampling cost and model performance.

\subsubsection{Comparison with Other Active Learning Methods}
Active learning, as an effective data selection strategy, can improve the generalization ability of models while reducing annotation costs. ALT is the first method to utilize active learning strategies to optimize the performance of DL compilers, while BALTO builds upon ALT by incorporating diversity features to further enhance the effectiveness of the active learning algorithm. Based on this, our method simultaneously considers three key factors: representativeness, diversity, and uncertainty, to further improve the sampling efficiency of DL compilers. As shown in Table \ref{tab4}, we compare the performance of our method with ALT and BALTO on CPU (E5-2673) and GPU (T4) hardware platforms at a 10$\%$ sampling rate. The results demonstrate that our method outperforms both ALT and BALTO in sampling effectiveness on the GPU platform T4 and the CPU platform E5-2673, exhibiting superior performance.

\begin{table}[!t]
  \centering
  \caption{Comparison of Sampling Methods on CPU and GPU Platforms}
  \begin{tabular}{l|cc|cc}
    \hline
     & \multicolumn{2}{c|}{T4} &  \multicolumn{2}{c}{E5-2673}\\
    \cline{2-5}
    & Top-1 Score & Top-5 Score & Top-1 Score& Top-5 Score \\
    \hline
    ALT   & 0.8668 & 0.9305 & 0.8521 & 0.9359 \\
    BALTO & 0.8466 & 0.8933 & 0.8619 & 0.9199 \\
    Ours   & \textbf{0.8761} & \textbf{0.9318} & \textbf{0.8806} & \textbf{0.9512} \\
    \hline
  \end{tabular}
  \label{tab4}
\end{table}

\subsection{Dataset-based Evaluation of CKD}
\subsubsection{Impact of Source Hardware Architecture on Target Hardware in CKD}
Different hardware platforms have distinct architectures, and there can be both knowledge-enhancing and knowledge-interfering effects between different hardware. To explore the mutual influence between different hardware during the CKD process, we conducted the experiments shown in Table \ref{tab5}. On the CPU side, we selected the i7-12700F as the target hardware and applied the RDU Sampler with a 10$\%$ sampling rate, while the remaining four CPU platforms served as source hardware. On the GPU side, the 3080Ti was used as the target hardware with the same 10$\%$ sampling rate, and the other two GPU platforms were used as source hardware.

As shown in Table \ref{tab5}, on the CPU hardware platform, we found that the knowledge from Platinum 8272CL and E5-2673 v4 facilitated the performance of i7-12700F, while the knowledge from EPYC 7452 and Graviton2 interfered with i7. This could be because Platinum 8272CL and E5-2673 v4 are both from Intel, while EPYC 7452 and Graviton2 are CPUs from AMD and ARM architectures, respectively. On the GPU hardware platform, the knowledge from T4 facilitated the performance of 3080Ti, while the knowledge from K80 interfered with it. This could be because both T4 and 3080Ti focus on high-performance computing, while K80 is more focused on inference tasks. \textcolor{black}{The experimental results show that, although the second component of the $L_{\text{KD}}$ loss in our CKD module constrains some harmful knowledge from the teacher model, it cannot fully prevent cross-hardware transfer of harmful knowledge.} Therefore, when using TCL, we try to perform knowledge transfer mainly between hardware from the same manufacturer to ensure the effectiveness of knowledge transfer.

\begin{table}[!t]
  \centering
  \caption{Impact of Different Source Hardware Platforms on Target Hardware Performance on CPU and GPU Platforms}
  \begin{tabular}{ccc|ccc}
    \hline
    \textbf{CPU} & Top-1 Score & Top-5 Score & \textbf{GPU} & Top-1 Score & Top-5 Score \\
    \hline
    i7-12700F & 0.9116 & 0.9753 & 3080Ti & 0.8533 & 0.8912 \\
    \hline
    \makecell[c]{i7-12700F\\Platinum 8272CL} & \textbf{0.9319} & \textbf{0.9754} & \makecell[c]{3080Ti\\T4} & \textbf{0.8675} & \textbf{0.9070} \\
    \hline
    \makecell[c]{i7-12700F\\Platinum 8272CL\\Graviton2} & 0.9067 & 0.9635 & \makecell[c]{3080Ti\\T4\\K80} & 0.8606 & 0.9031 \\
    \hline
    \makecell[c]{i7-12700F\\Platinum 8272CL\\Graviton2\\E5-2673 v4} & 0.9118 & 0.9660 & - & -  & -  \\
    \hline
    \makecell[c]{i7-12700F\\Platinum 8272CL\\Graviton2\\E5-2673 v4\\EPYC 7452} & 0.9031 & 0.9713 & -  & - & -  \\
    \hline
  \end{tabular}
  \label{tab5}
\end{table}

\begin{table}[!t]
  \centering
  \caption{Performance Comparison of Transfer Learning Methods on CPU and GPU Platforms}
  \begin{tabular}{lcc|cc}
    \toprule
    \multirow{2}{*}{} & \multicolumn{2}{c|}{\textbf{i7-12700F}} & \multicolumn{2}{c}{\textbf{3080Ti}} \\
    \cmidrule(lr){2-3} \cmidrule(lr){4-5}
    \textbf{Method} & Top-1 Score & Top-5 Score & Top-1 Score & Top-5 Score \\
    \midrule
    Fine-tuning & 0.8378 & 0.9182 & 0.7918 & 0.8599 \\
    MTL-TLP                  & 0.8744 & 0.9361 & 0.8228 & 0.8943 \\
    TCL                     & \textbf{0.8938} & \textbf{0.9366} & \textbf{0.8426} & \textbf{0.8997} \\
    \bottomrule
  \end{tabular}
  \label{tab6}
\end{table}

\subsubsection{Comparison with Other Transfer Learning Methods}
Transfer learning aims to transfer existing knowledge to new but related tasks, thereby reducing the reliance on large-scale data and computational resources in the target domain and improving learning efficiency and generalization capability. To validate the effectiveness of our proposed cross-hardware transfer learning approach, we compare the performance of fine-tuning, multi-task learning, and our CKD method on both CPU and GPU platforms.

For the CPU experiments, we use the Intel i7-12700F as the target hardware platform and randomly sample 10$\%$ of the dataset for training, with the Intel Platinum 8272CL platform serving as the source hardware. For the GPU experiments, the NVIDIA 3080Ti is used as the target hardware, with 10$\%$ of the dataset randomly sampled, and the NVIDIA T4 platform used as the source hardware. The detailed results are presented in Table \ref{tab6}.

As shown in Table \ref{tab6}, our proposed TCL-based continual distillation method outperforms fine-tuning and multi-task learning on both the CPU platform (Intel i7-12700F) and the GPU platform (NVIDIA 3080Ti). The fine-tuning approach only uses the pre-trained model from the source platform (Intel Platinum 8272CL) as the initial weights and retrains on the i7-12700F platform, resulting in poor performance and failing to achieve knowledge transfer across multiple hardware platforms. Multi-task learning shows improved performance; however, it relies on simultaneous input of data from both the source and target hardware, and as the number of hardware platforms increases, the model parameters grow rapidly, leading to scalability issues. In contrast, our CKD method does not cause model parameter growth as the number of hardware platforms increases. It effectively leverages all existing hardware knowledge and is better suited for multi-hardware platform transfer tasks in DL compiler scenarios.

\begin{figure}[!t]
  \centering
  \begin{subfigure}[b]{0.49\textwidth}
    \centering
    \includegraphics[width=\textwidth]{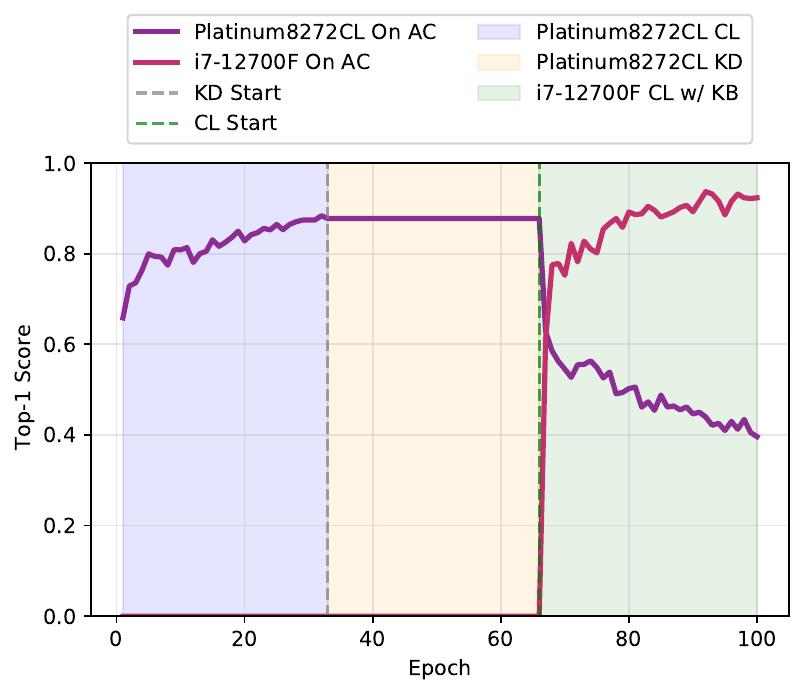}
    \caption{CPU Platforms (i7-12700F and Platinum8272CL).}
    \label{fig:a1}
  \end{subfigure}
  \begin{subfigure}[b]{0.49\textwidth}
    \centering
    \includegraphics[width=\textwidth]{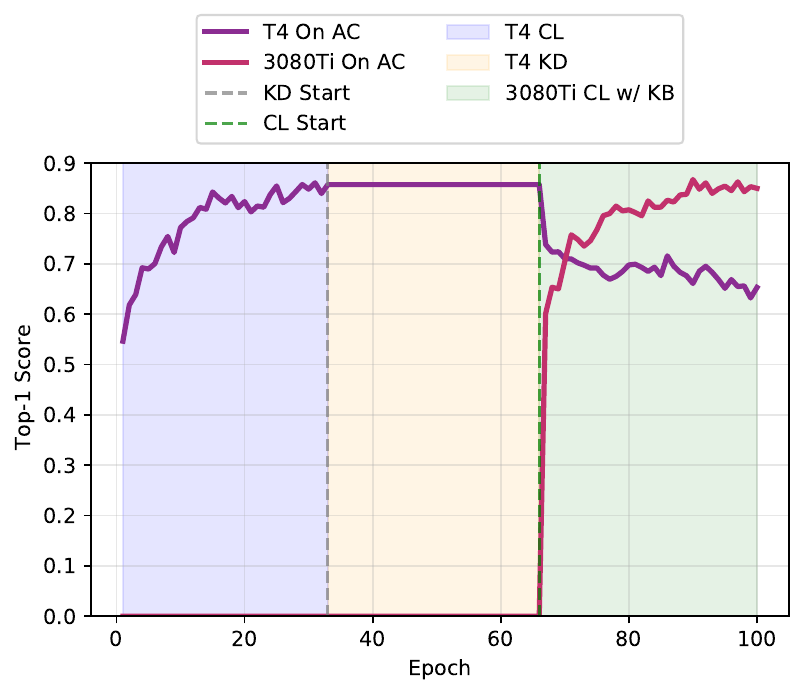}
    \caption{GPU Platforms (3080Ti and T4)}
    \label{fig:b1}
  \end{subfigure}
  \caption{The Top-1 Score of Alternating CL and KD on CPU/GPU.}
  \label{fig8}
\end{figure}

\subsubsection{Top-1 score Iteration in CKD Alternating Training}
In our CKD framework, the system consists of two components: KB and AC. KB stores knowledge from different hardware platforms, while AC continuously learns new knowledge and distills it back into KB. Within the CKD framework, learning follows a “source platform first, target platform last” order, leveraging multi-source knowledge in KB to enhance learning on the target platform. Once continual learning on the target platform is complete, no further KD is applied, as the objective has been achieved. As shown in Fig. \ref{fig8}, we present Top-1 score curves across epochs during alternating CL and KD phases on AC for both CPU (i7-12700F and Platinum8272CL) and GPU (3080Ti and T4) platforms.

On the CPU platform (Fig. \ref{fig:a1}), during Platinum8272CL’s CL phase, Top-1 score on AC rises and gradually stabilizes, while i7-12700F has not yet started learning (Top-1 score = 0). In the KD phase of Platinum8272CL, AC’s parameters are frozen, and its Top-1 score remains unchanged; i7-12700F still has a Top-1 score of 0. During i7-12700F’s CL phase, knowledge from KB is utilized, leading to a steady increase in its accuracy, while AC updates cause a drop in Platinum8272CL’s Top-1 score. The GPU platform shows a similar pattern (Fig. \ref{fig:b1}). During T4’s CL phase, Top-1 score on AC rises and remains stable during KD. Before 3080Ti’s CL phase, its Top-1 score is 0. During 3080Ti’s CL phase, its Top-1 score steadily increases, while T4’s accuracy slightly decreases, likely due to the similar architectures of the two GPUs.




\subsection{Ablation Study}
To verify the effectiveness of each module in our design for the DL compiler, we conducted systematic ablation studies on TCL. Specifically, we individually added the RDU sampler, the Mamba architecture, and the KCD module to evaluate the impact of each module on the target hardware platforms Intel i7 and NVIDIA 3080Ti. The baseline method used was the current mainstream DL compiler TLP, and the cost model was trained with 10$\%$ randomly sampled data. The experimental results are detailed in Table \ref{tab8}.

The experimental data clearly show that all three key modules play a crucial role in improving model performance. First, the RDU sampler significantly reduces data collection time through an intelligent sampling strategy, thereby enhancing overall training efficiency. Second, the model based on the Mamba architecture excels at capturing the intrinsic dependencies within scheduling primitive sequences, modeling the complexity of hardware scheduling more accurately than traditional architectures. Finally, the KCD module enables knowledge collaboration across hardware platforms, allowing TCL to better adapt to diverse hardware environments. The combination of these three modules enables our designed DL compiler TCL to achieve significantly improved prediction accuracy and hardware adaptability while maintaining efficient training.

\begin{table}[!t]
  \centering
  \caption{Ablation Study of TCL on CPU and GPU Platforms}
  \begin{tabular}{ccc|cc|cc}
    \toprule
    \multirow{2}{*}{\makecell[c]{RDU}} & \multirow{2}{*}{\makecell[c]{Mamba}} & \multirow{2}{*}{\makecell[c]{CKD}} 
    & \multicolumn{2}{c|}{\textbf{i7-12700F}} 
    & \multicolumn{2}{c}{\textbf{3080Ti}} \\
    \cmidrule(lr){4-5} \cmidrule(lr){6-7}
    & & & Top-1 Score & Top-5 Score & Top-1 Score & Top-5 Score \\
    \midrule
    \XSolidBrush & \XSolidBrush & \XSolidBrush & 0.7414 & 0.9337 & 0.6614 & 0.7613 \\
    \Checkmark & \XSolidBrush & \XSolidBrush & 0.9020 & 0.9733 & 0.8349 & 0.8673 \\
    \Checkmark & \Checkmark & \XSolidBrush & 0.9116 & \textbf{0.9753} & 0.8533 & 0.8912 \\
    \Checkmark & \Checkmark & \Checkmark & \textbf{0.9319} & 0.9713 & \textbf{0.8675} & \textbf{0.9031} \\
    \bottomrule
  \end{tabular}
  \label{tab8}
\end{table}

\subsection{End-to-end Evaluation}
The end-to-end evaluation aims to comprehensively assess the overall performance of TCL in optimizing DL models on different hardware platforms after its integration into the auto-tuning framework Ansor and the DL compiler TVM in real-world scenarios. Specifically, the evaluation focuses on two key aspects: (1) the tuning time required to achieve the same inference latency as the baseline model under the same number of tuning iterations, reflecting the speed and efficiency of the tuning process; and (2) the final inference latency of the optimized model on the target hardware platform given a fixed number of tuning iterations, demonstrating the actual improvement in performance optimization. To ensure the representativeness and broad applicability of the evaluation, we selected five representative DL models and conducted systematic tests with 2000 tuning iterations on both CPU and GPU, the two mainstream hardware platforms. We compared TCL with Ansor (zero-shot cost model), Felix\cite{zhao2024felix} (few-shot cost model, GPU only), and the fully supervised cost-model approaches Tenset-MLP and MTL-TLP. \textcolor{black}{Additionally, we evaluated MetaTune and One-Shot Tuner; however, MetaTune is not open-source and cannot be fairly reproduced, and One-Shot Tuner, based on AutoTVM, is limited by its search space and performs significantly worse than Ansor, and is therefore excluded from the comparison.}

\begin{figure}[!t]
\centering
\includegraphics[width=\linewidth]{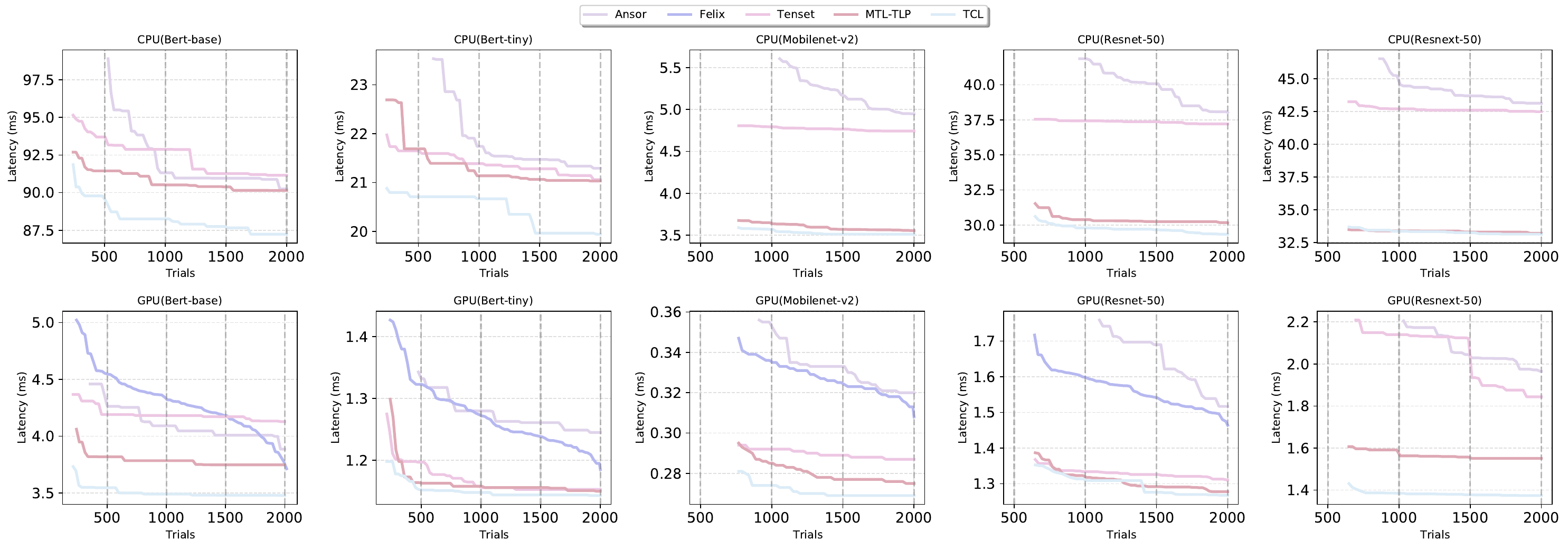}
\caption{Tuning Curves of Five Representative Models on CPU and GPU.}
\label{fig9}
\end{figure}

\subsubsection{Tuning Curve Analysis of Representative DL Models on CPU and GPU}
We compared different tuning methods on CPU and GPU platforms across multiple iterations, shown as tuning curves in Fig. \ref{fig9}. To ensure reliable results, the minimum iterations for each model were set to the number of subgraphs $\times 24$, so curves do not start from zero. Some initial Ansor values were removed for clarity, as its early performance was poor with high latency. Felix supports only GPU and cannot handle ResNeXt-50 due to group convolutions.

Ansor and Felix show large curve fluctuations on both platforms because Ansor relies on online data and Felix samples data randomly, requiring many iterations to converge. In contrast, Tenset and MTL-TLP use sufficient offline data, leading to smaller fluctuations and faster convergence. TCL, though trained offline with limited data, uses the RDU Sampler to collect more representative and diverse data, improving cost model training. On the Bert-Base model, Ansor and Felix achieve better final tuning results than Tenset, showing that large-scale datasets do not always generalize well across models.

\subsubsection{Comparison of Fully Supervised Cost-Model-Based Tuning Performance on CPU and GPU}

\begin{figure}[!t]
  \centering

  \begin{subfigure}[b]{0.48\textwidth}
    \centering
    \includegraphics[width=\textwidth]{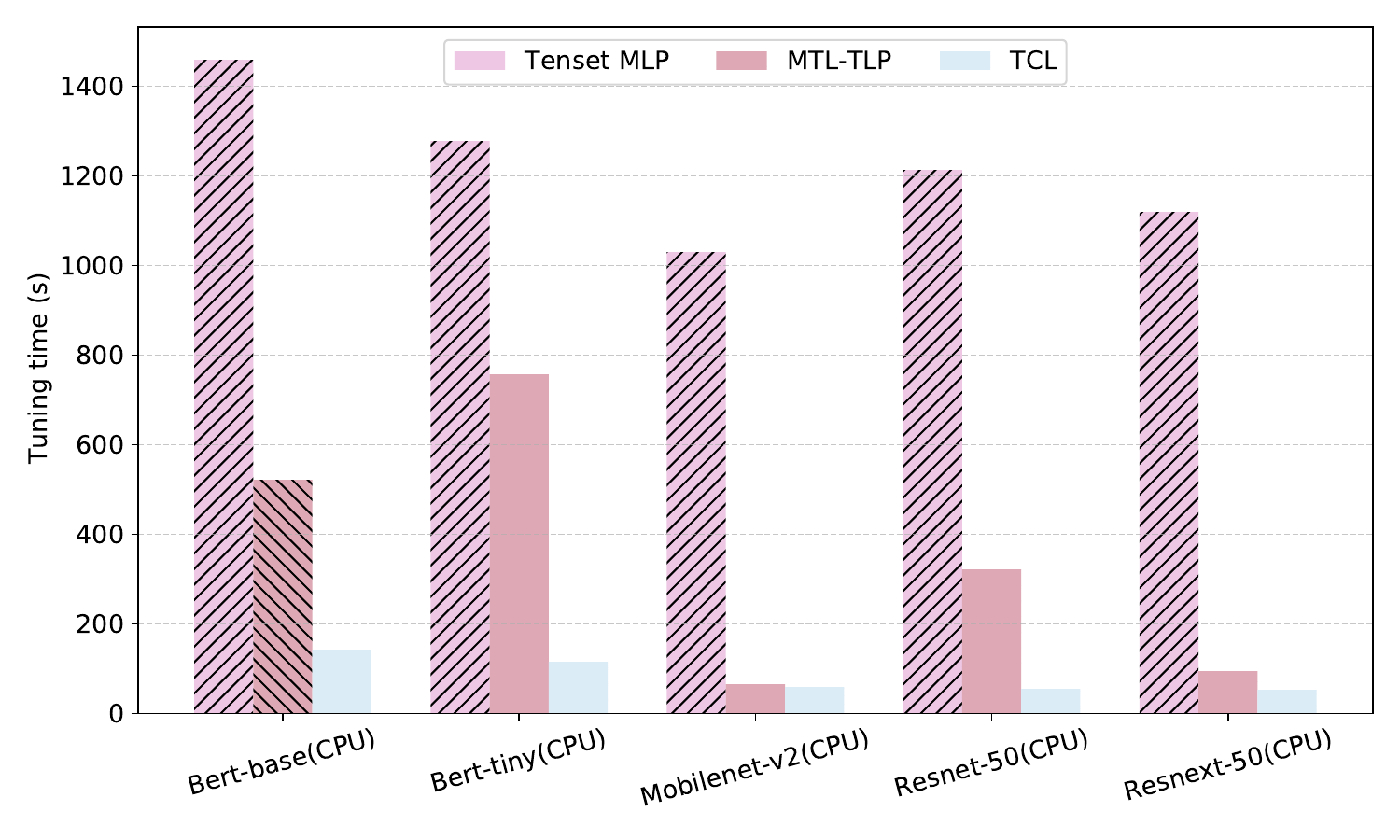}
    \caption{CPU (i7-12700F): Tuning Time to Match Tenset-MLP Inference Latency Across Models}
    \label{CPU tuning time1}
  \end{subfigure}
  \hspace{0.02\textwidth}
  \begin{subfigure}[b]{0.48\textwidth}
    \centering
    \includegraphics[width=\textwidth]{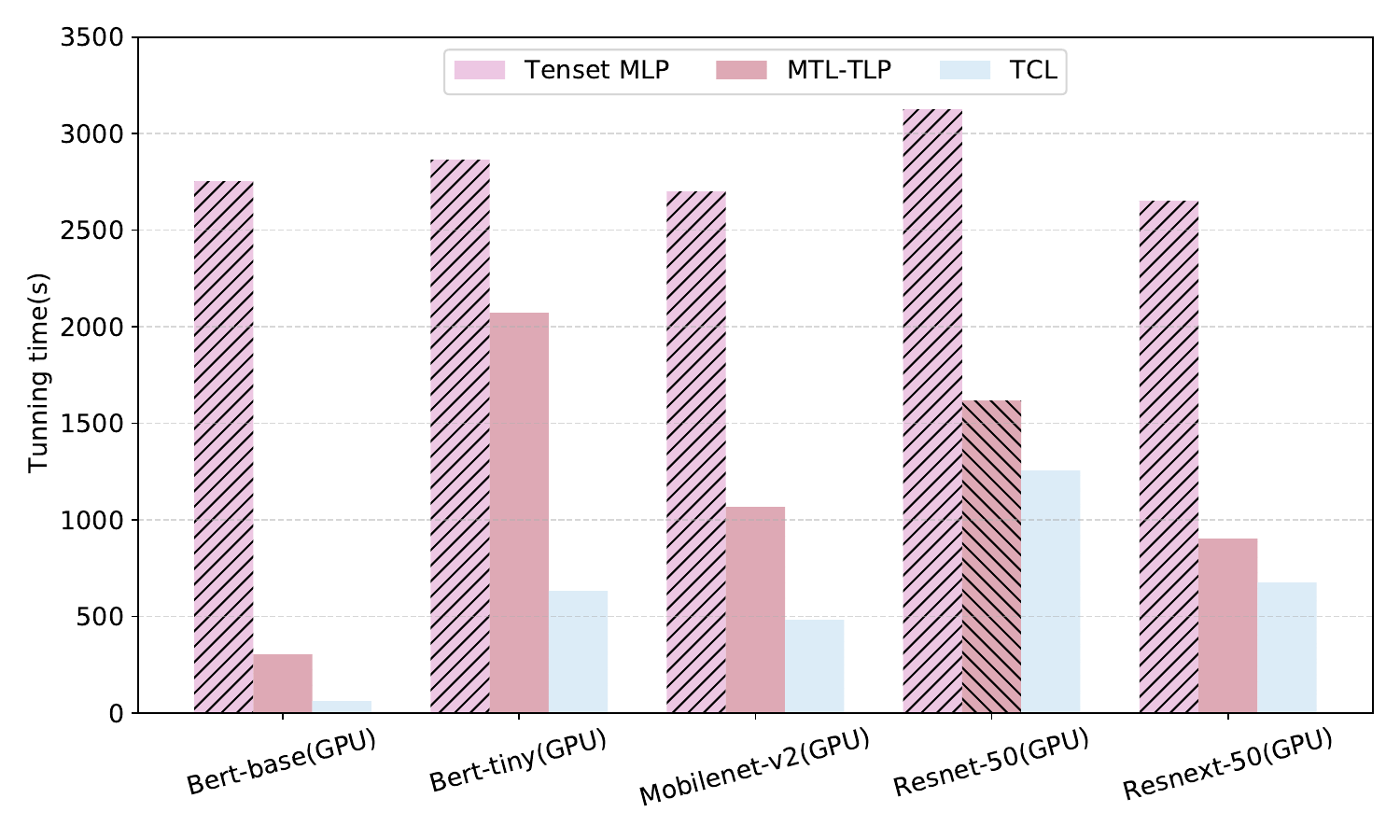}
    \caption{GPU (RTX 3080Ti): Tuning Time to Match Tenset-MLP Inference Latency Across Models}
    \label{GPU tuning time1}
  \end{subfigure}

  \caption{Comparison of Tuning Time Required to Match Tenset-MLP's Inference Latency Across Different Models}
  \label{fig10}
\end{figure}

\begin{figure}[!t]
  \centering

  \begin{subfigure}[b]{0.48\textwidth}
    \centering
    \includegraphics[width=\textwidth]{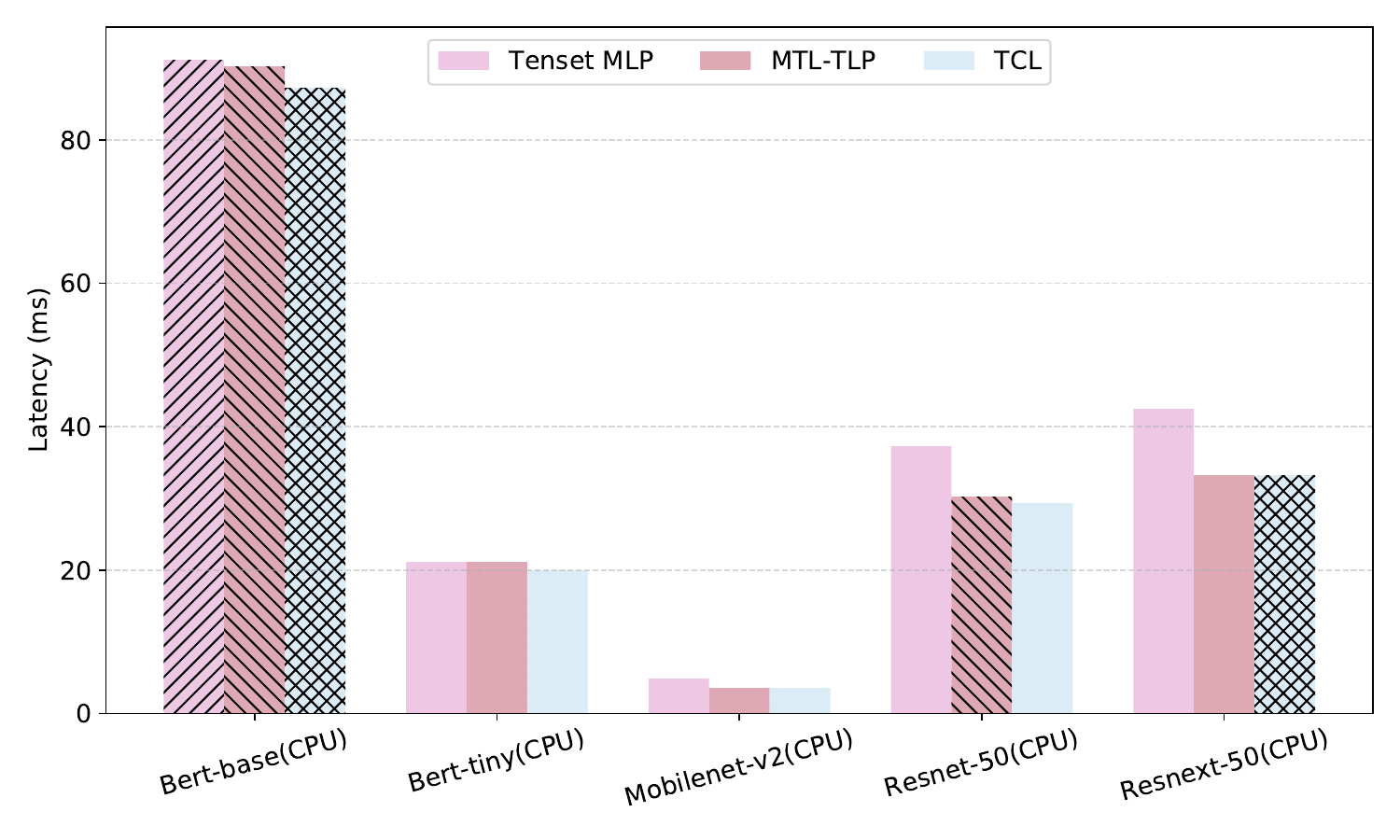}
    \caption{CPU (i7-12700F): Inference Latency with 2000 Tuning Trials}
    \label{CPU lantency1}
  \end{subfigure}
  \hspace{0.02\textwidth}
  \begin{subfigure}[b]{0.48\textwidth}
    \centering
    \includegraphics[width=\textwidth]{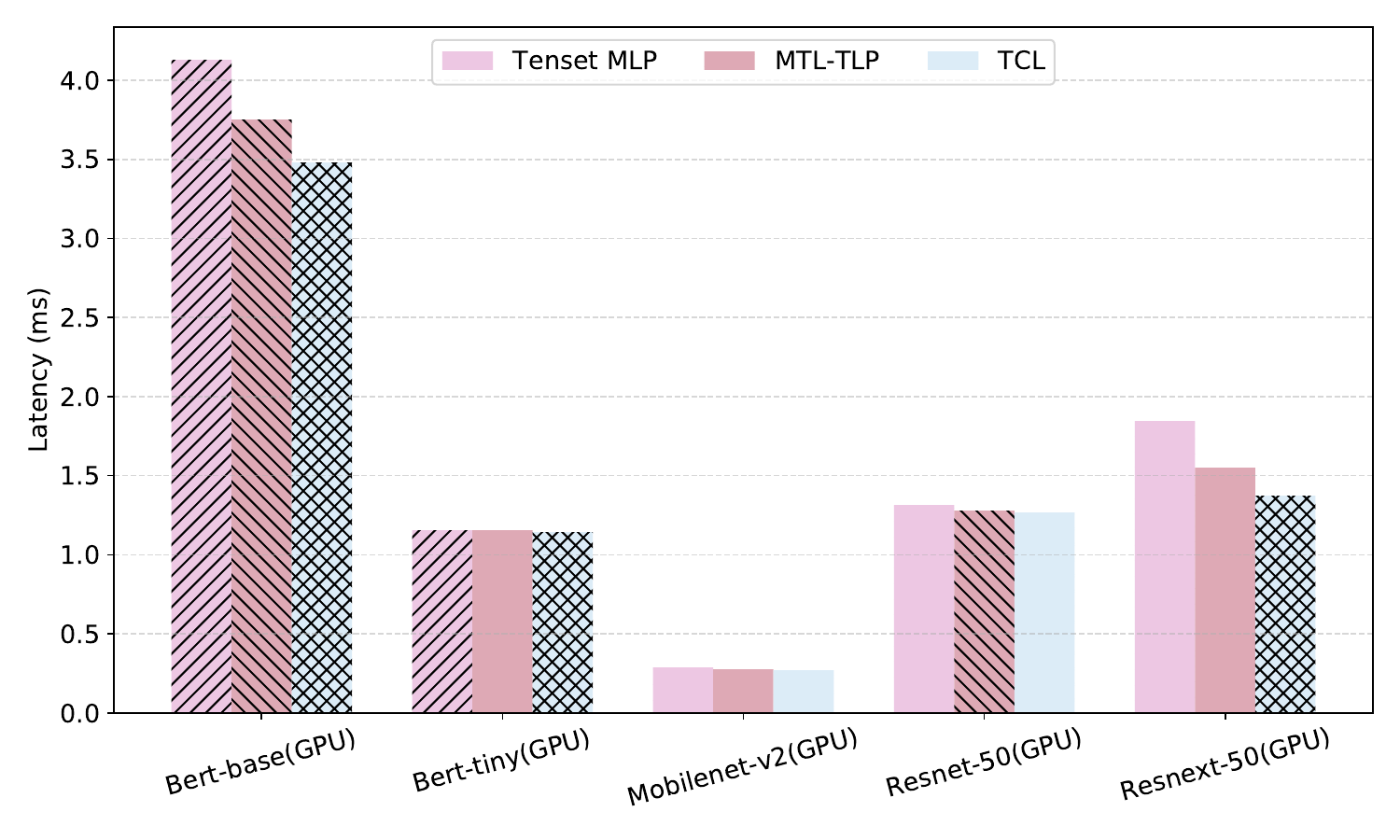}
    \caption{GPU (RTX 3080Ti): Inference Latency with 2000 Tuning Trials}
    \label{GPU lantency1}
  \end{subfigure}

  \caption{Inference Latency Achieved Under 2000 Tuning Trials Across Different Models}
  \label{fig11}
\end{figure}

We compared the tuning time required by TCL and MTL-TLP to reach the same inference latency that Tenset-MLP achieves after 2000 tuning iterations (Fig. \ref{fig10}). On CPUs, TCL delivers an average tuning-time speedup of 16.8$\times$ over Tenset-MLP and 3.84$\times$ over MTL-TLP. On GPUs, the speedups are 12.48$\times$ and 2.63$\times$, respectively. We also compared the inference latency of Tenset-MLP, MTL-TLP, and TCL under the same 2000 iterations (Fig. \ref{fig11}). On CPUs, TCL improves latency by 1.20$\times$ over Tenset-MLP and 1.03$\times$ over MTL-TLP; on GPUs, the improvements are 1.13$\times$ and 1.05$\times$, respectively.

Overall, TCL delivers much larger tuning-efficiency gains on CPUs than on GPUs. This is because CPUs have higher baseline latency and thus greater optimization space, while GPUs’ stronger compute capability and lower baseline latency limit possible improvements. Although TCL, MTL-TLP, and Tenset-MLP all use offline-trained cost models, MTL-TLP leverages multi-task learning to improve prediction accuracy across platforms. Building on this, TCL more effectively utilizes cross-platform knowledge while substantially reducing MTL-TLP’s parameter size, resulting in significantly improved tuning efficiency.

\subsubsection{Comparison of Zero-Shot and Few-Shot Cost-Model-Based Tuning Performance on CPU and GPU}
\begin{figure}[!t]
  \centering

  \begin{subfigure}[b]{0.48\textwidth}
    \centering
    \includegraphics[width=\textwidth]{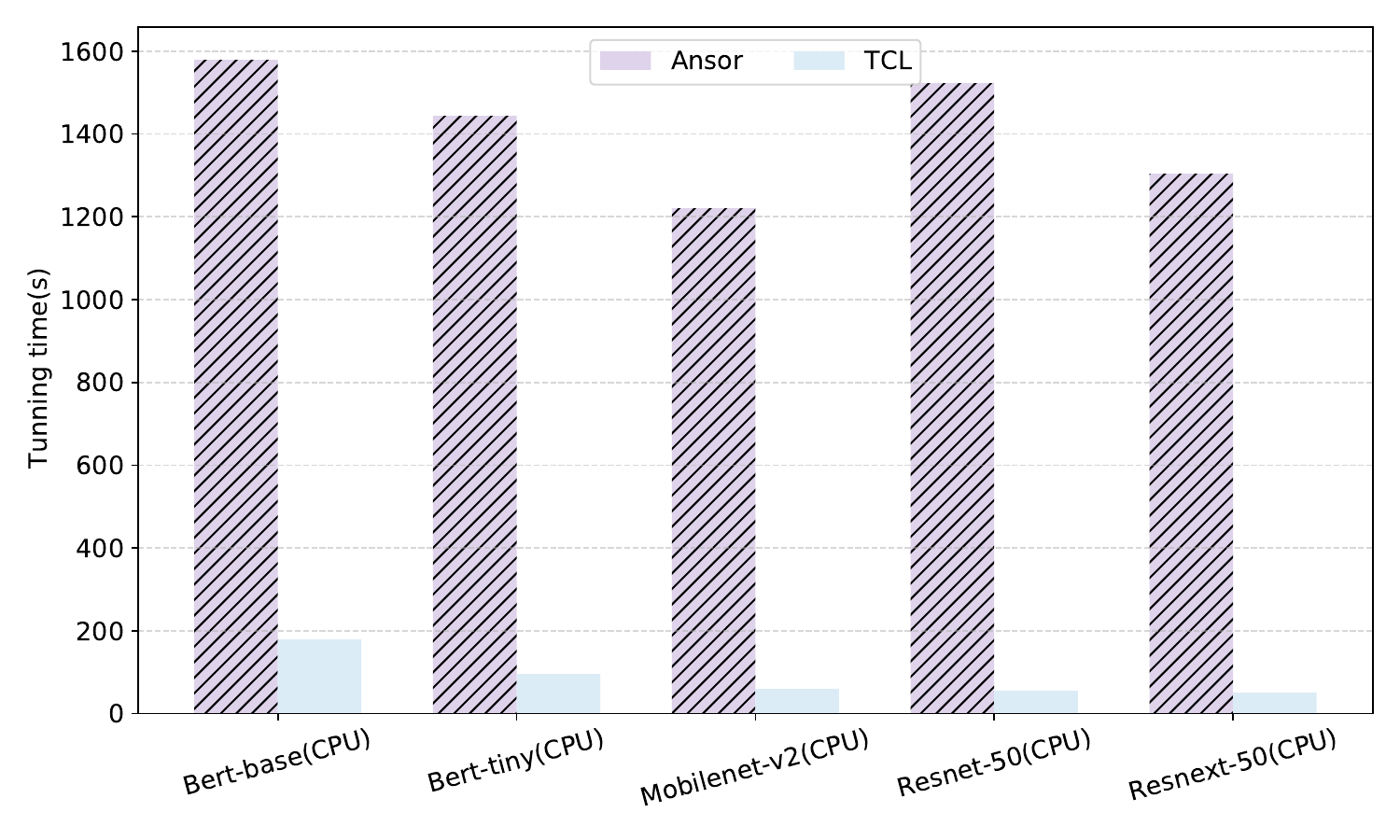}
    \caption{CPU (i7-12700F): Tuning Time to Match Ansor Inference Latency Across Models}
    \label{CPU tuning2}
  \end{subfigure}
  \hspace{0.02\textwidth}
  \begin{subfigure}[b]{0.48\textwidth}
    \centering
    \includegraphics[width=\textwidth]{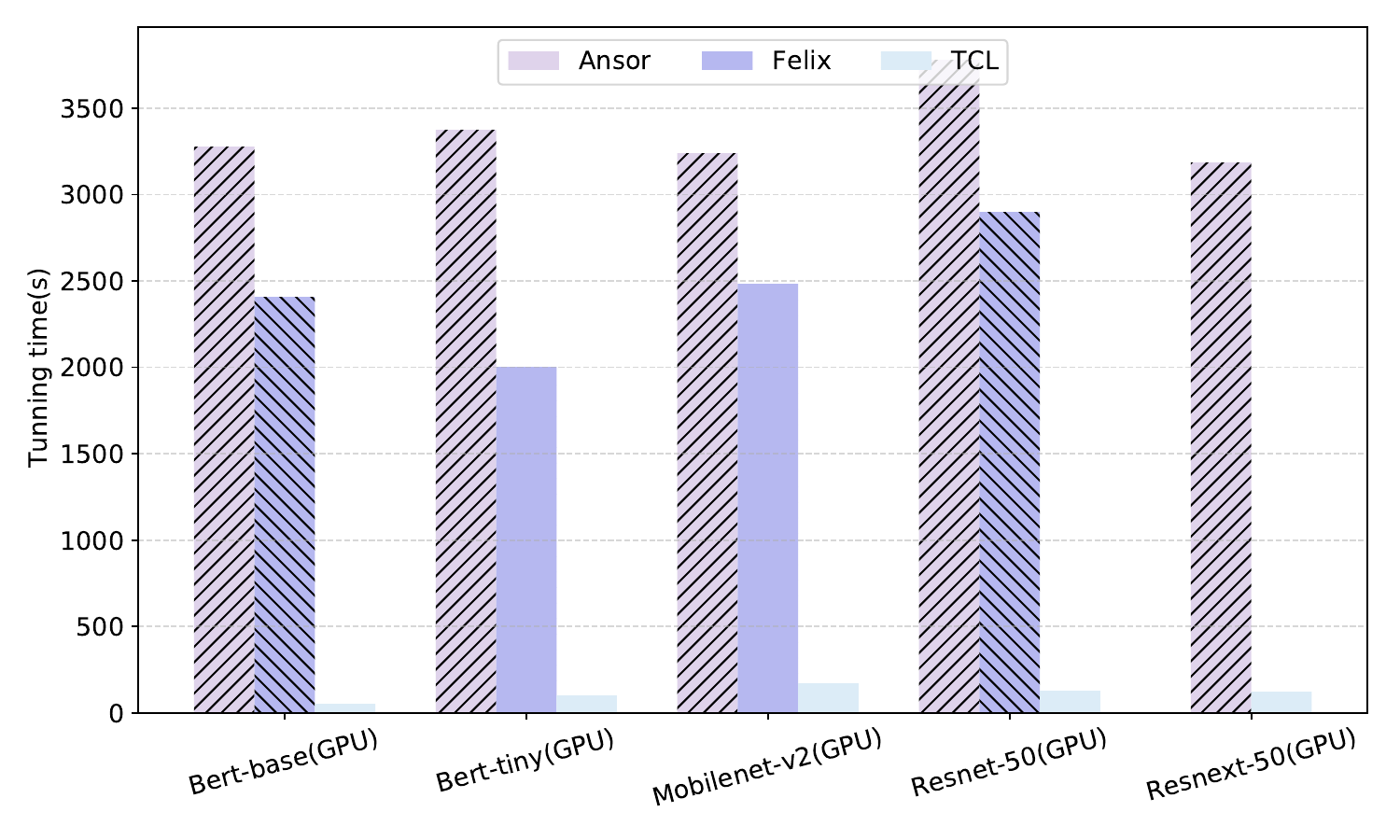}
    \caption{GPU (RTX 3080Ti): Tuning Time to Match Ansor Inference Latency Across Models}
    \label{GPU tuning2}
  \end{subfigure}

  \caption{Comparison of Tuning Time Required to Match Ansor's Inference Latency Across Different Models.}
  \label{fig12}
\end{figure}

We compare the tuning time required by TCL and Felix to achieve the same inference latency as Ansor after 2000 tuning iterations, as shown in Fig. \ref{fig12}. On the CPU platform, TCL achieves an average speedup of 19.98$\times$ over Ansor; on the GPU platform, TCL achieves 34.78$\times$ speedup over Ansor and 26.23$\times$ over Felix.

\begin{figure}[!t]
  \centering

  \begin{subfigure}[b]{0.48\textwidth}
    \centering
    \includegraphics[width=\textwidth]{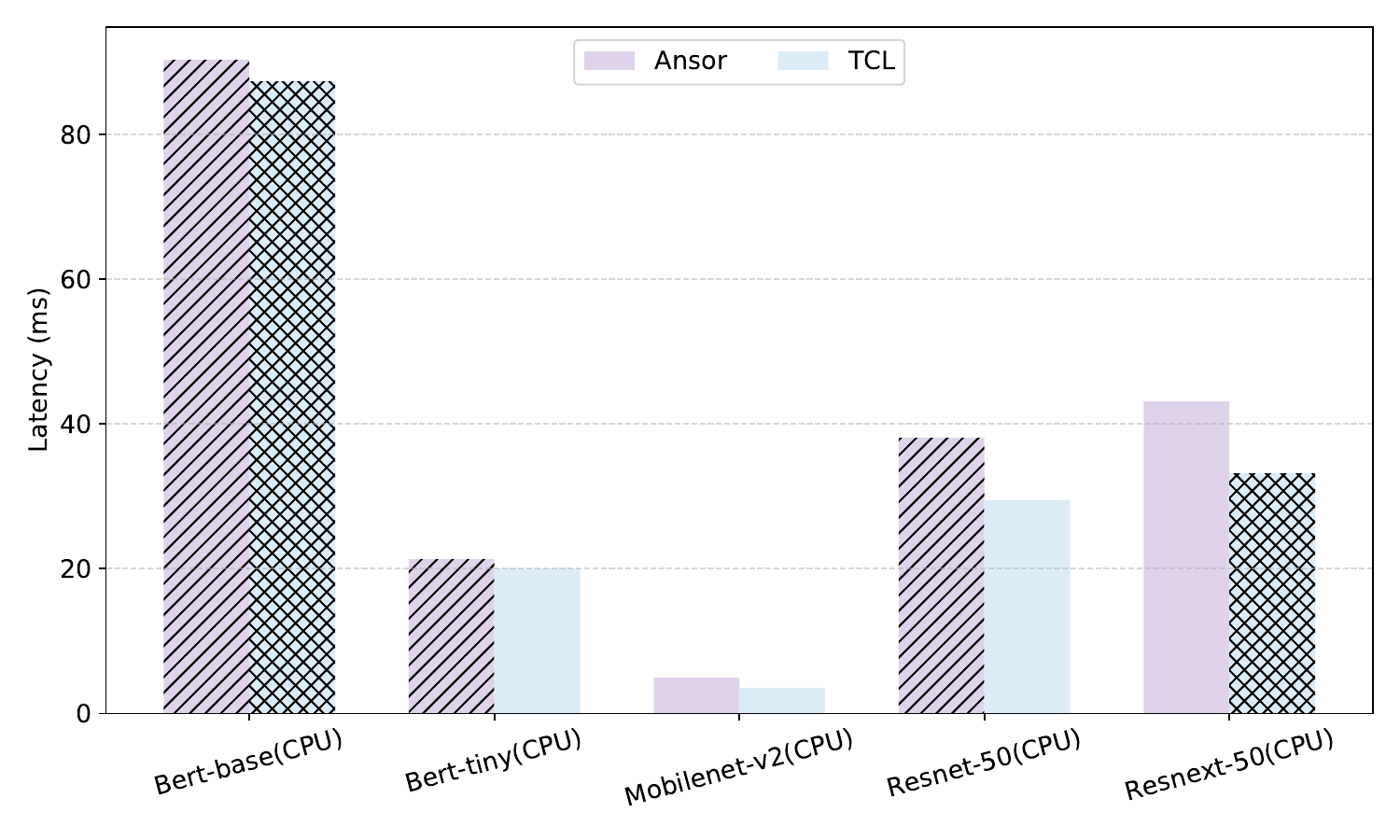}
    \caption{CPU (i7-12700F): Inference Latency with 2000 Tuning Trials}
    \label{CPU latency2}
  \end{subfigure}
  \hspace{0.02\textwidth}
  \begin{subfigure}[b]{0.48\textwidth}
    \centering
    \includegraphics[width=\textwidth]{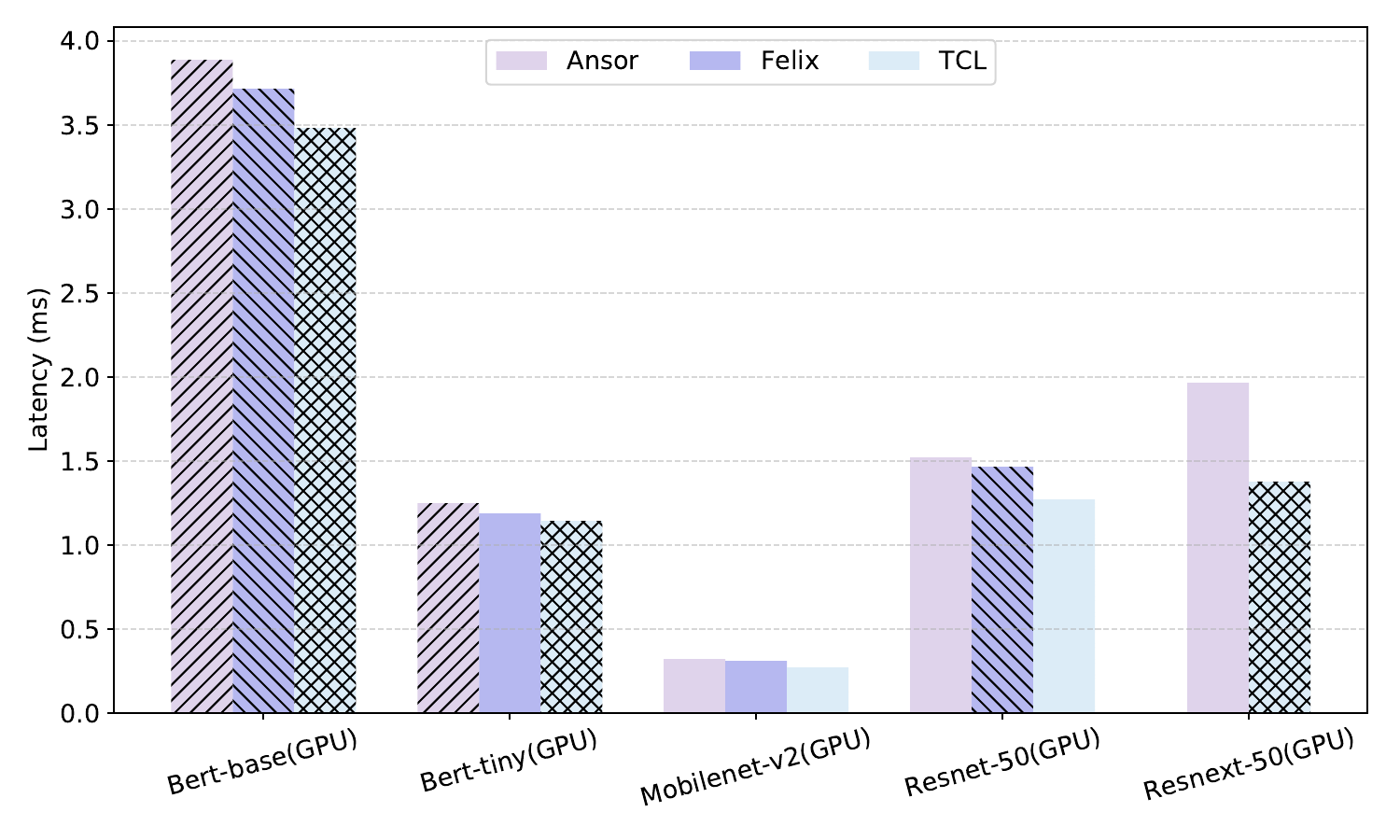}
    \caption{GPU (RTX 3080Ti): Inference Latency with 2000 Tuning Trials}
    \label{GPU latency2}
  \end{subfigure}

  \caption{Inference Latency Achieved Under 2000 Tuning Trials Across Different Models.}
  \label{fig13}
\end{figure}

We further compare the inference latency of models optimized by Ansor, Felix, and TCL under the same 2000 tuning iterations, as shown in Fig. \ref{fig13}. On the CPU platform, TCL improves average inference latency by 1.22$\times$ over Ansor; on the GPU platform, TCL improves inference latency by 1.21$\times$ over Ansor and 1.10$\times$ over Felix. Overall, compared with the zero-shot method Ansor and the few-shot method Felix, TCL achieves significant improvements in tuning efficiency. This is mainly because the cost models of Ansor and Felix must be trained online during the search process; insufficient training results in weaker prediction performance, making it difficult for them to accurately identify high-performance tensor programs. In contrast, TCL employs a fully offline-trained cost model, enabling more stable and efficient guidance during tuning.

%% file: 8_Conclusion.tex
\section{Conclusion}
This paper proposes TCL, which aims to address the challenges faced by current mainstream DL compilers based on offline-trained cost models during cross-hardware migration. By introducing an efficient data collection strategy, named RDU Sampler, TCL achieves comparable prediction performance using only 10$\%$ of the original data on the target device. Meanwhile, a lightweight Mamba architecture is adopted to reconstruct the cost model, effectively capturing the relationships among schedule primitive sequences and further improving prediction accuracy. In addition, a CKD framework is designed to efficiently integrate knowledge from multiple source devices. This mitigates the issues of parameter explosion and data dependency that commonly arise in conventional multi-task learning-based migration methods when the number of source devices increases, enhancing compiler's adaptability. The TCL-based DL compiler not only significantly reduces data collection time on the target hardware, but also achieves strong performance in both tuning time and inference latency. Although there is still room for improvement in inference latency optimization, future work will focus on designing more fine-grained cost models to further improve prediction accuracy and enhance inference performance.

\begin{acks}
This work was supported by the Big Data Computing Center of Southeast University, the Key R\&D Program of Guangdong Province (Project No.~2021B1101270006), Shandong Provincial Natural Science Foundation (No.~ZR2023QF056), and the Jiangsu Provincial Science and Technology Major Special Project (Grant No.~BG2024032).
\end{acks}
